\documentclass[lettersize,journal]{IEEEtran}
\usepackage{amsmath,amsfonts}
\usepackage{algorithmic}
\usepackage{algorithm}
\usepackage{array}
\usepackage[caption=false,font=normalsize,labelfont=sf,textfont=sf]{subfig}
\usepackage{textcomp}
\usepackage{stfloats}
\usepackage{url}
\usepackage{verbatim}
\usepackage{graphicx}
\usepackage{cite}

\usepackage{xcolor}
\usepackage{bm}
\usepackage{booktabs}
\usepackage{multirow}
\usepackage{pgfplots}
\usepackage{makecell}
\usepackage{hyperref}

\begin{document}

\title{Neural Network Optimization Reimagined: Decoupled Techniques for Scratch and Fine-Tuning}
\author{Xin Ning\textsuperscript{*},~\IEEEmembership{Senior Member,~IEEE}, Qiankun Li\textsuperscript{*},~\IEEEmembership{Member,~IEEE},
		  Xiaolong Huang,
            Qiupu Chen,
            Feng He, \\
            Weijun Li,~\IEEEmembership{Senior Member,~IEEE},
            Prayag Tiwari,~\IEEEmembership{Senior Member,~IEEE},
		Xinwang Liu\textsuperscript{\dag},~\IEEEmembership{Senior Member,~IEEE}
		
		\thanks{This work is funded by Beijing Natural Science Foundation (No. L233036) and National Natural Science Foundation of China (No. 62373343). }
            \thanks{Xin Ning and Qiankun Li contributed equally to this work}
            \thanks{Xin Ning and Weijun Li are with Institute of Semiconductors, Chinese Academy of Sciences, Beijing, 10083, China.}
            \thanks{Qiankun Li is with the College of Computing and Data Science (CCDS), Nanyang Technological University, 639798, Singapore.}
		\thanks{Qiupu Chen and Feng He are both with the University of Science and Technology of China, Hefei 230027, China.}
            \thanks{Xiaolong Huang is with Mila-Quebec AI Institute, Montreal, QC H2S 3H1, Canada.}
            \thanks{Prayag Tiwari is with the School
of Information Technology, Halmstad University, Sweden.}
		\thanks{Xinwang Liu is with College of Computer, National University of Defense Technology, Changsha, 410073, China.}
		\thanks{Xinwang Liu is the corresponding author (Email: xinwangliu@nudt.edu.cn).}
	}

% \markboth{IEEE Transactions on Pattern Analysis and Machine Intelligence}%
% 	{Shell \MakeLowercase{\textit{et al.}}: Bare Demo of IEEEtran.cls for IEEE Journals}
	

\markboth{IEEE Transactions on Pattern Analysis and Machine Intelligence,~Vol.~xx, No.~xx, Month~202x}%
{Ning \MakeLowercase{\textit{et al.}}: Neural Network Optimization Reimagined: Decoupled Techniques for Scratch and Fine-Tuning}


\maketitle

\begin{abstract}
With the accumulation of resources in the era of big data and the rise of pre-trained models in deep learning, optimizing neural networks for various tasks often involves different strategies for fine-tuning pre-trained models versus training from scratch. 
However, existing optimizers primarily focus on reducing the loss function by updating model parameters, without fully addressing the unique demands of these two major paradigms. In this paper, we propose DualOpt, a novel approach that decouples optimization techniques specifically tailored for these distinct training scenarios. For training from scratch, we introduce real-time layer-wise weight decay, designed to enhance both convergence and generalization by aligning with the characteristics of weight updates and network architecture. For more importantly fine-tuning, we integrate weight rollback with the optimizer, incorporating a rollback term into each weight update step. This ensures consistency in the weight distribution between upstream and downstream models, effectively mitigating knowledge forgetting and improving fine-tuning performance. Additionally, we extend the layer-wise weight decay to dynamically adjust the rollback levels across layers, adapting to the varying demands of different downstream tasks. Extensive experiments across diverse tasks, including image classification, object detection, semantic segmentation, and instance segmentation, demonstrate the broad applicability and state-of-the-art performance of DualOpt.
Code is available at https://github.com/qklee-lz/OLOR-AAAI-2024.
\end{abstract}


\section{Introduction}
\IEEEPARstart{E}{arly} deep learning models were typically trained from scratch, and model parameters were initialized randomly \cite{li2023embracing, huang2024one}. However, with the rapid growth of data and advances in deep learning \cite{laion-2b,imagenet-21k,laion-400m}, a wealth of publicly available pre-trained models has emerged \cite{openclip, mae, beit}. These pre-trained models have significantly enhanced performance in various tasks through transfer learning and fine-tuning techniques \cite{transfer-pami1, transfer-pami2}. Consequently, fine-tuning has become the mainstream approach in many deep learning applications \cite{li2023data, li2024advancing, zou2025rhythmformer}, though training from scratch is still essential for certain tasks and specific scenarios \cite{strach-pami1, strach-pami2}.

Traditional neural network optimizers focus primarily on minimizing the loss function by updating model parameters \cite{xie2021positive, kosson2023rotational}. In addition, weight decay is often combined with optimization algorithms to constrain the model weights, helping to promote better generalization \cite{hanson1988comparing, ghiasi2024improving}. While these methods are effective, they often overlook the distinct needs of training paradigms. Training from scratch requires careful balancing to ensure efficient convergence and robust generalization, whereas fine-tuning faces the challenge of knowledge forgetting, where the model may lose valuable information learned during pre-training \cite{de2021continual}.

\begin{figure}[t!]
    \centering
    \includegraphics[width=0.45\textwidth]{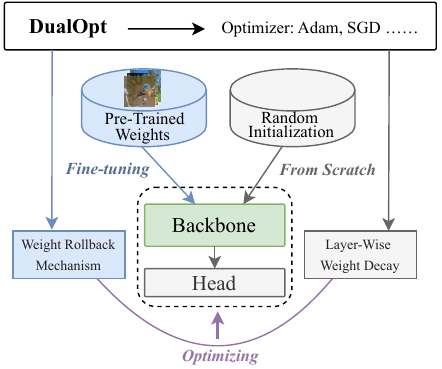}
    %\caption{Overview of the proposed DualOpt. It integrates weight rollback for fine-tuning and layer-wise weight decay for training from scratch, offering flexibility and efficiency across tasks.}
    \caption{Overview of the proposed DualOpt. The optimization is both fine-tuning with pre-trained weights and training from scratch with random initialization. The DualOpt integrates a weight rollback mechanism for fine-tuning and layer-wise weight decay for training from scratch, providing flexibility and efficiency for various tasks.}
    \label{fig:overview}
\end{figure}

In the deep learning domain, optimizers like Stochastic Gradient Descent (SGD) \cite{sgd} and Adaptive Moment Estimation (Adam) \cite{adam} are widely used. Researchers have continually proposed improvements to enhance optimization efficiency \cite{ergen2023globally, stock2019equi}, focusing on gradient propagation \cite{mom}, parameter update mechanisms \cite{duchi2011adaptive}, and regularization techniques \cite{krogh1991simple}. 
Momentum was introduced to accelerate convergence by incorporating past gradient information \cite{mom}. Adaptive Gradient assigned individual learning rates to parameters \cite{duchi2011adaptive}, a concept further refined in the Adam optimizer \cite{adam}. Weight decay was introduced to prevent overfitting and improve generalization \cite{krogh1991simple}, prompting a series of widely used regularization techniques \cite{cnnsgdvitadam, AdamW, AdamW_2}.
However, these methods often overlook the structural characteristics of neural networks. When training from scratch, shallow layers capture low-level features (e.g., color, edges, texture), with less risk of overfitting, while deeper layers focus on high-level features (e.g., objects, semantics), which are more prone to overfitting. 
On the other hand, fine-tuning requires more preserving the general shallow layer information acquired during the pre-training phase.
Therefore, the same decay rate across the entire network results in a coarse decay effect.
In addition, existing fine-tuning methods have not fully considered the design of optimizers. Replay-based methods \cite{r_16, r_49, replay2020, replay2022} focus on incorporating pre-trained data, but this approach is often inefficient. EWC \cite{EWC} and L2-SP \cite{L2-SP} use regularization to constrain weight parameters, but they are even incompatible with adaptive optimizers \cite{adam, AdamW, AdamW_2}. Parameter isolation methods \cite{VPT, VPT_cvpr}, which tailor new branches or modules for downstream tasks and networks, introduce additional training parameters and generally lack versatility.

%In this paper, we propose DualOpt, a novel optimizer that reimagines neural network optimization by decoupling the strategies for fine-tuning and training from scratch. For training from scratch, DualOpt introduces a real-time layer-wise weight decay mechanism, whcih dynamically adjusts the weight decay across different layers of the network. This approach enhances the network's convergence rate and generalization ability by aligning the optimization process with the structural and functional characteristics of the network. 

%For fine-tuning, DualOpt introduces a weight rollback term to the weight update term, allowing the model to gradually approach the pre-trained weights while learning the downstream task, thereby mitigating knowledge forgetting and enhancing fine-tuning performance. 
%In addition, a extended layer-wise penalty is devised to employ penalty decay and the diversified decay rate to adjust the weight rollback levels of layers. Penalty decay combines feature pyramids with transfer learning, giving more significant weight rollback to shallow features such as color and texture, and smaller weight backtracking to deep features such as semantic information. The diversified decay rate according to the similarity between the two domain data to adjust the level of weight rollback. Finally, our method enables the fine-tuning model to extract more generalized features from shallow layers and adapt varying downstream task objectives.
In this paper, we propose DualOpt, a novel optimizer that reimagines neural network optimization by decoupling the strategies for fine-tuning and training from scratch. The overview is shown in Fig. \ref{fig:overview}. For training from scratch, DualOpt introduces a real-time layer-wise weight decay mechanism that dynamically adjusts the decay rate across different layers of the network. This method enhances convergence speed and generalization by tailoring the optimization process to the structural and functional characteristics of the network.

For fine-tuning, DualOpt introduces a weight rollback term to the weight update term, allowing the model to gradually approach the pre-trained weights while adapting to downstream tasks. This mitigates knowledge forgetting and improves fine-tuning performance. 
In addition, we extend this approach by employing a layer-wise penalty mechanism with penalty decay and a diversified decay rate. Penalty decay integrates feature pyramids and transfer learning, applying more significant rollback to shallow layers that capture features like color and texture, and less rollback to deeper layers that focus on semantic features. The diversified decay rate adjusts rollback levels based on the similarity between the pre-training and downstream task domains, enabling more adaptive and task-specific fine-tuning.

%Through comprehensive experiments on tasks including image classification, object detection, semantic segmentation, and instance segmentation, we demonstrate that DualOpt not only achieves state-of-the-art performance but also offers a generalizable solution applicable to a wide range of neural network training scenarios.

We validate the effectiveness of our proposed DualOpt through extensive experiments on tasks including image classification \cite{cifar-100, svhn, cub-200, standfordcars, places-lt, ip102, officehome, PACS}, object detection \cite{coco}, semantic segmentation \cite{ade20k}, and instance segmentation \cite{coco} on 10 popular datasets. The results demonstrate that DualOpt achieves state-of-the-art performance and offers a generalizable solution for a wide range of training scenarios.

The contributions of this paper are summarized as follows:
\begin{itemize}
\item We propose DualOpt, an optimizer that decouples strategies for training from scratch and fine-tuning to address distinct needs of these two major training paradigms.
\item A real-time layer-wise weight decay mechanism is introduced to enhance convergence and generalization during training from scratch.
\item For fine-tuning, we propose a weight rollback mechanism combined with layer-wise penalty decay, mitigating knowledge forgetting and improving task adaptation.
%\item The proposed DualOpt achieves state-of-the-art performance on ten popular datasets across various tasks, including various types of image classification, multiple pre-trained models, and image detection and segmentation.
\item The proposed DualOpt achieves state-of-the-art performance on ten popular datasets across various tasks.
\end{itemize}

This work is an extension of our earlier conference version presented at AAAI 2024 \cite{huang2024one}. Compared to the conference version, the journal version includes the following key enhancements:
\begin{itemize}
\item The previous fine-tuning optimizer is significantly improved with detailed theoretical proofs, efficiency analyses, validation experiments, and comprehensive ablation studies.
\item We expand the original work into a unified framework by developing DualOpt to effectively address both fine-tuning and training from scratch, making it more generalizable and versatile.
\item To evaluate the effectiveness of the newly proposed DualOpt, extensive experiments are conducted on 10 popular datasets across diverse tasks, including image classification (In-Distribution, Out-of-Distribution, and large-scale datasets), object detection, semantic segmentation, and instance segmentation.
\end{itemize}

%This work is an extension of our earlier conference version presented at AAAI 2024 \cite{huang2024one}. Here, we significantly expand the scope of the original work by not only enhancing the fine-tuning process but also addressing the challenges of training from scratch. These improvements lead to the development of a unified and more generalizable DualOpt optimization framework. In addition, we provide a more thorough theoretical analysis of the optimizer, and conduct extensive experiments on additional datasets to evaluate the newly proposed mechanisms and methods. Moreover, we further conduct complementary validation experiments, analyze detailed results, and explore more ablation studies.

The structure of this paper is as follows. Section \ref{sec:RelatedWork} reviews existing optimizer design and fine-tuning work and briefly clarifies the superiority of our method. Section \ref{sec:method} presents the architectural design of the proposed method. 
The experimental configuration are outlined in Section \ref{sec:experiment}. Section \ref{sec:results_analysis} includes the results and discussion that demonstrate the effectiveness of the proposed method. Finally, Section \ref{sec:conclusion} concludes our work.

\section{Related Work} %预训练资源，数据集、模型; 微调方法；优化器
\label{sec:RelatedWork}
\subsection{Training from Scratch, Pre-training and Fine-tuning}

In early deep learning models, training was predominantly conducted from scratch, requiring large datasets and computational resources \cite{lecun2015deep}. As models and datasets grew in size, this approach became increasingly resource-intensive \cite{brown2020language}. 

Currently, the deep learning field has numerous pre-trained resources available, including large-scale datasets and pre-trained models. These datasets cover many domains, such as ImageNet \cite{deng2009imagenet}, COCO \cite{coco}, and Open Images \cite{kuznetsova2020open}, which have been widely applied in tasks like image classification, object detection, and image segmentation. In the natural language processing domain, large-scale corpora such as Common Crawl \cite{dodge2021documenting} and BooksCorpus \cite{zhu2015aligning} provide rich linguistic information for pre-training. Based on these datasets, researchers have developed several high-performance pre-trained models, such as BERT \cite{devlin2018bert}, GPT \cite{brown2020language}, CLIP \cite{openclip}, and T5 \cite{T5}, which have become foundational for many applications, demonstrating exceptional performance across various tasks \cite{zou2025rhythmformer, zou2025rhythmmamba}.
Therefore, pre-trained models, trained on large-scale datasets have since emerged as a more efficient alternative, enabling models to learn general representations and later adapt through fine-tuning \cite{devlin2018bert}. 
However, despite being a powerful approach, fine-tuning cannot completely eliminate the need for training from scratch in certain specific scenarios \cite{bommasani2021opportunities}.

One of the challenges in fine-tuning pre-trained models is catastrophic forgetting. To address this, existing methods can be broadly categorized into three main types: replay methods, regularization methods, and parameter isolation methods. Replay methods involve periodically retraining on portions of upstream task data to help the model maintain memory of previous tasks and balance new and old information \cite{wu2018incremental}. These methods can somewhat mitigate catastrophic forgetting. For instance, some earlier works employed memory-based mechanisms to store portions of upstream task data and periodically replay them for training \cite{hsu2018re}. However, as the scale of pre-trained models and datasets increases, the cost of storing and managing this data becomes substantial, especially when dealing with contemporary large-scale pre-trained models \cite{brown2020language}. Regularization methods impose constraints during parameter updates to reduce damage to the original knowledge in the fine-tuning process. For example, the Fisher information matrix can be used to limit drastic changes in model parameters \cite{zenke2017continual}, or weight decay and L2 regularization can control parameter changes \cite{li2017learning}. Although these methods can effectively reduce catastrophic forgetting in certain scenarios, they usually rely on existing optimizers for parameter updates rather than being specifically designed for fine-tuning tasks. Moreover, regularization methods may not perform ideally when dealing with complex downstream tasks \cite{EWC, rannen2017encoder}. Parameter isolation methods involve introducing independent modules or branches into the pre-trained model to avoid directly modifying the core parameters of the pre-trained model. These methods use adapter modules or low-rank decomposition modules to adapt to downstream tasks \cite{houlsby2019parameter, hu2021lora}. The advantage of this approach is that it preserves the original model structure, but it also introduces additional training parameters, increasing the model's complexity. Parameter isolation methods typically require multiple rounds of freezing and unfreezing strategies, which adds to the complexity of the training process and the difficulty of hyperparameter tuning \cite{pfeiffer2020adapterfusion}.

Although these methods partially address catastrophic forgetting, most of them focus on model architecture and data processing rather than exploring the problem from the optimizer’s perspective. 
Different from the existing works, this paper proposes a novel optimization framework that introduces real-time layer-wise weight decay for improving convergence when training from scratch and a weight rollback mechanism for mitigating catastrophic forgetting during fine-tuning. By dynamically adjusting optimization strategies for each training paradigm, DualOpt enhances both performance and efficiency.

\subsection{Optimizers}
Optimizers play a critical role in training deep neural networks, as their primary task is to iteratively update the model’s parameters to minimize the loss function. %The design of the optimizer directly impacts the model’s convergence speed, stability, and generalization ability. In recent years, many classical optimization algorithms have been widely applied and have demonstrated excellent performance across various tasks. 
Stochastic Gradient Descent (SGD) is one of the earliest and most widely adopted optimization methods. By using gradient estimates from a small random sample in each iteration, it makes the training process more efficient and avoids the high computational costs associated with some global optimization algorithms \cite{robbins1951stochastic}. Despite its simplicity and effectiveness in practice, SGD often relies on manually adjusting the learning rate and can easily get stuck in local minima or saddle points when optimizing non-convex objectives. Momentum-based gradient descent introduces a momentum term in the update direction, which effectively reduces gradient oscillations and accelerates convergence. Nesterov Accelerated Gradient (NAG) \cite{sutskever2013importance} further optimizes momentum methods by anticipating the gradient direction and making adjustments in advance, thereby improving convergence efficiency. While momentum methods address some limitations of SGD, the introduction of adaptive optimizers further enhanced optimization efficiency. Adaptive Gradient Method (AdaGrad) \cite{duchi2011adaptive} is an early adaptive optimization algorithm that introduces different learning rates for each parameter, allowing for more efficient updates, especially when handling sparse data. However, a major issue with AdaGrad is that the learning rate continuously decreases during training, causing the update steps to become too small in later stages. To address this problem, Adam (Adaptive Moment Estimation) \cite{adam} was introduced. Adam combines the advantages of momentum and adaptive learning rates by tracking both the first-order and second-order moments (i.e., the mean and squared gradients) simultaneously, and it has shown strong performance in practical applications. Because Adam adaptively adjusts the learning rate for each parameter, it performs exceptionally well in handling complex deep neural networks and has become the default optimizer in many tasks. However, Adam also has some drawbacks, such as potentially leading to overfitting in some tasks, and instability during training in certain cases. 
By adjusting the learning rate warm-up period, RAdam \cite{liu2019variance} avoids over-adjustment in the early stages of training, improving Adam’s convergence. AdaBound \cite{luo2019adaptive} combines the advantages of SGD and Adam, using Adam's adaptive learning rate in the early stages of training and then gradually transitioning to SGD to prevent instability later in the process.
Recent robust fine-tuning methods~\cite{ghiasi2024improving,tian2024rethinking} revisit weight decay by introducing adaptive or schedule-based decay coefficients tailored for foundation models.

However, existing optimizers often overlook the unique needs of the two different training paradigms: training from scratch and fine-tuning. In training from scratch, model parameters are randomly initialized, and the optimizer’s primary task is to ensure fast convergence while avoiding overfitting. In the fine-tuning process, the model has already learned some general knowledge from pre-training, so the optimizer needs to avoid over-adjusting the pre-trained weights, thereby mitigating the problem of catastrophic forgetting. 
In this direction, we propose DualOpt, an optimizer specifically designed for these two training paradigms. Unlike existing optimizers, DualOpt introduces a real-time layer-wise weight decay mechanism to improve convergence when training from scratch, and a weight rollback mechanism to reduce catastrophic forgetting during fine-tuning.

\begin{figure*}[t]
	\centering 
        \includegraphics[width=0.9\linewidth]{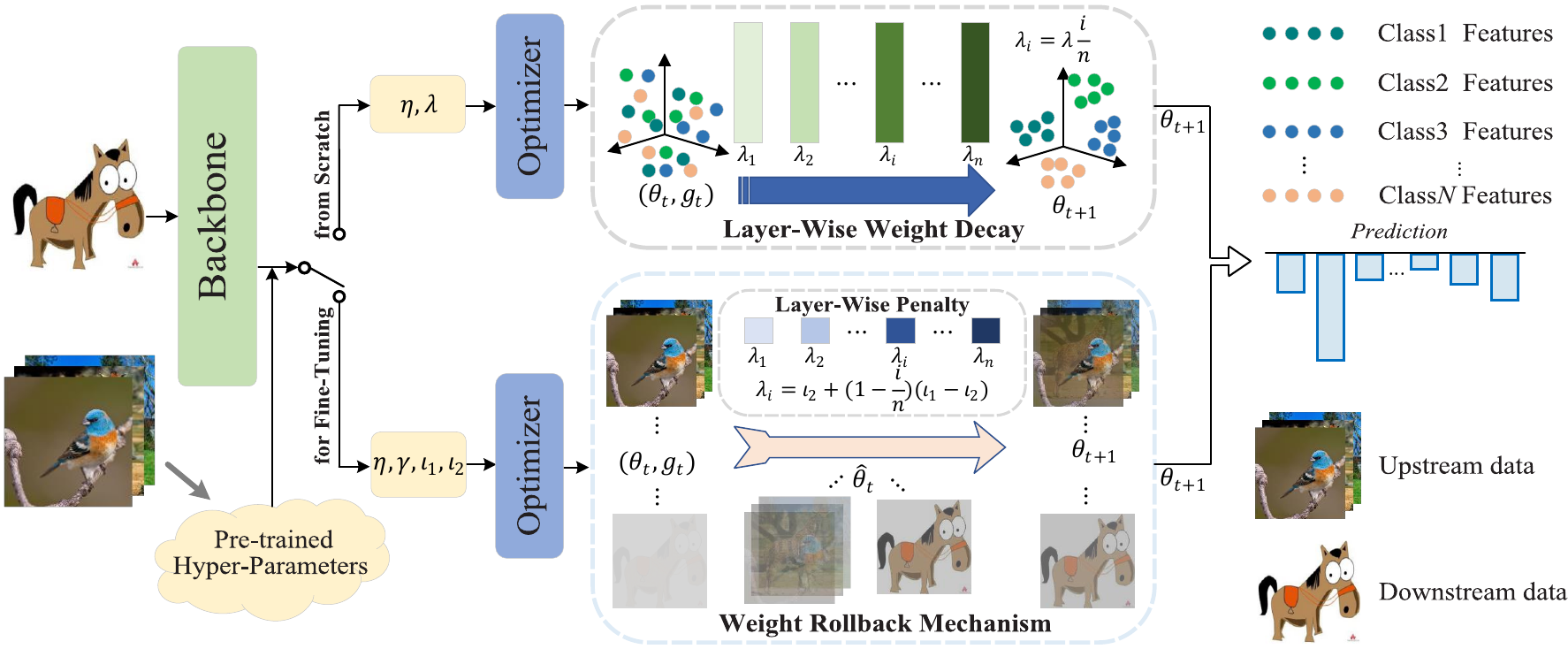}
	\caption{Overview of the proposed DualOpt optimization framework. The framework addresses training from scratch with real-time layer-wise weight decay (top pathway), and fine-tuning with weight rollback and layer-wise penalty (bottom pathway). The weight rollback mechanism ensures alignment with pre-trained weights, while the layer-wise penalty adapts the learning intensity across layers based on depth, reducing knowledge forgetting. In addition, popular optimizers such as SGD and Adam are incorporated into DualOpt.} 
	\label{fig_1}
\end{figure*}

\section{Method}
\label{sec:method}
In this section, we illustrate DualOpt, a novel optimization framework designed to address the distinct needs of both training from scratch and fine-tuning. The overall architecture is shown in Fig. \ref{fig_1}.  DualOpt decouples these two training paradigms by incorporating a real-time layer-wise weight decay mechanism for training from scratch and a combination of weight rollback and layer-wise penalty decay for fine-tuning. The proposed method can incorporate popular optimizers such as SGD and Adam, providing flexibility and robustness across different neural networks and tasks, ensuring improved performance.

\subsection{Real-Time Layer-Wise Weight Decay for From Scratch}
\label{method:FS}
%\subsection{Real-Time Layer-Wise Weight Decay for Training from Scratch}

Optimizing neural networks from scratch requires careful consideration of both convergence speed and generalization. Traditional optimizers like SGD and Adam incorporate weight decay as a regularization method to prevent overfitting by penalizing large weight magnitudes. 
The weight $\theta$ updating at time step $t$ can be simplified as:
\begin{equation}
    \label{eq1}
    \theta_{t+1} = \theta_t - \eta_t U^g_t - \lambda \theta_t,
\end{equation}
where $\lambda$ represents the weight decay rate, and $U^g_t$ is the update term calculated from the gradient, then multiplied by the learning rate $\eta_t$. 
However, the application of weight decay typically lags behind the parameter update, introducing a delay that may reduce the regularization's effectiveness. 
Specifically, when the condition $(\theta_t - \eta_t U^g_t - \lambda \theta_t)^2 > (\theta_t - \eta_t U^g_t)^2$ holds, the weight value tends to increase in magnitude, which is contrary to the expected regularization effect. Theoretically, this situation is possible, as proven below:

Firstly, expanding the inequality:
\begin{equation}
(\theta_t - \eta_t U^g_t)^2 - 2\lambda \theta_t (\theta_t - \eta_t U^g_t) + \lambda^2 \theta_t^2 > (\theta_t - \eta_t U^g_t)^2.
\end{equation}
Then, eliminating identical terms yields:
\begin{equation}
-2\lambda \theta_t (\theta_t - \eta_t U^g_t) + \lambda^2 \theta_t^2 > 0.
\end{equation}
Rearranging the terms, we get:
\begin{equation}
-2\lambda \theta_t (\theta_t - \eta_t U^g_t) > - \lambda^2 \theta_t^2.
\end{equation}
Finally, dividing both sides by $-2\lambda \theta_t$ (assuming $\theta_t \neq 0$ and $\lambda \neq 0$) yields:
\begin{equation}
\theta_t - \eta_t U^g_t < \frac{\lambda}{2} \theta_t,
\end{equation}
which leads to two cases:
\begin{equation}
\begin{cases}
	\eta_t U^g_t < \left( 1 - \frac{\lambda}{2} \right) \theta_t, & \text{if } \theta_t < 0, \\
	\eta_t U^g_t > \left( 1 - \frac{\lambda}{2} \right) \theta_t, & \text{if } \theta_t > 0.
\end{cases}
\label{eq:conditions}
\end{equation}
This indicates that under certain conditions as Eq. (\ref{eq:conditions}), the application of weight decay can lead to weight increases, countering the expected regularization effect. 
%Similarly, this issue can be observed in other regularization optimization techniques, such as $L_2$ and $L_1$ regularization.

To address the above question, we simultaneously decays both the current weights and the update term. This can be expressed as
\begin{equation}
\theta_{t+1} = \theta_t - \eta_t U^g_t - \lambda (\theta_t - \eta_t U^g_t).
\end{equation}
It guarantees that as \(\lambda \to 1\), the weights \(\theta_t\) are driven towards zero, thereby enhancing the stability and generalization of the model. The proof is as follows:

First, expand the left-hand side to get:
\begin{equation}
(\theta_t - \eta_t U^g_t - \lambda (\theta_t-\eta_t U^g_t))^2 = (\theta_t - \eta_t U^g_t)^2(1-\lambda)^2.
\end{equation}
The right-hand side is:
\begin{equation}
(\theta_t - \eta_t U^g_t)^2.
\end{equation}
Thus, the inequality transforms into:
\begin{equation}
(\theta_t - \eta_t U^g_t)^2(1-\lambda)^2 \leq (\theta_t - \eta_t U^g_t)^2.
\end{equation}
Since $(\theta_t - \eta_t U^g_t)^2$ is non-negative, we can safely divide by it to get:
\begin{equation}
(1 - \lambda)^2 \leq 1.
\end{equation}
Therefore, the inequality holds for all $\lambda \in (0, 1]$.

In addition to real-time application, we further enhance this mechanism by making the decay rate layer-specific. Deep neural networks have a hierarchical structure, where shallow layers capture low-level features (e.g., edges, textures) with a lower risk of overfitting, and deeper layers capture high-level abstract features (e.g., semantics, object categories) that are more prone to overfitting. To reflect this, we propose a layer-wise weight decay rate:
\begin{equation}
\label{eq:lamda}
    \lambda_i = \lambda \frac{i}{n},
\end{equation}
\begin{equation}
\theta_{t+1} = \theta_t - \eta_t U^g_t - \{ \lambda_i  \theta_t^{(i)}\}_{i=1}^n(\theta_t - \eta_t U^g_t),
\end{equation}
where \( \lambda_i \) is the decay rate for layer \( i \), \( \lambda \) is the base decay coefficient, and \( n \) is the total number of layers. This design ensures a gradual increase in the decay rate as we move deeper into the network, providing stronger regularization for deeper layers while allowing more flexibility for the shallow layers.
Finally, Algorithm \ref{from-strach} illustrates $\text{DualOpt}_{Scratch}$, which integrates real-time weight decay and allows customization of the decay rate for each layer based on its network depth, easily adaptable to various optimizers.

\begin{algorithm}[tb!]
	\caption{Real-Time Layer-Wise Weight Decay for Training from Scratch}
	\footnotesize
	\label{from-strach}
	\begin{algorithmic}[1]
		\STATE{\textbf{input}: $\eta \in \mathbb{R}$: learning rate, $\lambda \in \mathbb{R}$: base weight decay factor, $n$: number of layers}
		\STATE{\textbf{initialize}: $t \leftarrow 0$,
		layer-wise decay factors $\lambda_i = \lambda \cdot \frac{i}{n}$ for $i=1,\dots,n$}
		\REPEAT
			\STATE{$t \leftarrow t + 1$}
			\STATE{$g_t \leftarrow \nabla_\theta f_t(\theta_{t})$ \hfill(Get batch gradient)}
			\STATE{$U^g_t \leftarrow g_t$}
			\STATE{$\hat{\theta}_{t+1} \leftarrow \theta_{t} - \eta_{t} U^g_t$ \hfill(Gradient update)}
			\STATE{$\theta_{t+1} \leftarrow \hat{\theta}_{t+1} - \lambda_i \hat{\theta}_{t+1}$ \hfill(Apply real-time layer-wise weight decay)}
		\UNTIL{stopping criterion is met}
		\RETURN{optimized parameters $\theta_{t+1}$}
	\end{algorithmic}
\end{algorithm}

\subsection{Weight Rollback Mechanism for Fine-Tuning}

The proposed weight rollback mechanism is a dynamic regularization approach designed to enhance fine-tuning by aligning current model weights with pre-trained weights. This mechanism operates in real-time, closely following each weight update to facilitate knowledge retention during the adaptation process.

The first step calculates the temporary weight \(\theta_{\text{pre}}\) based on the model weights from the previous time step:
\begin{equation}
	\label{eq3}
	\theta_{\text{pre}} = \theta_{t} - \eta_t g_t,
\end{equation}
where \(\theta_{t}\) represents the model weights from the previous step, \(\eta_t\) is the learning rate at the current time step, and \(g_t\) denotes the gradient.

Next, we compute the discrepancy \(\Delta d\) between \(\theta_{\text{pre}}\) and the pre-trained weights \(\theta_0\):
\begin{equation}
	\label{eq4}
	\Delta d = \theta_{\text{pre}} - \theta_0.
\end{equation}
This discrepancy is then incorporated into the weight update process, leading to the adjusted model weights \(\theta_{t+1}\):
\begin{equation}
	\label{eq5}
	\theta_{t+1} = \theta_{t} - \eta_t g_t - \lambda \Delta d.
\end{equation}
Substituting Eq. (\ref{eq3}) and Eq. (\ref{eq4}) into Eq. (\ref{eq5}) gives:
\begin{equation}
	\label{eq6}
	\theta_{t+1} = (1 - \lambda)(\theta_{t} - \eta_t g_t) + \lambda \theta_0.
\end{equation}
This formulation ensures that as \(\lambda\) approaches 1, the weights \(\theta_{t+1}\) converge to the pre-trained weights \(\theta_0\), effectively maintaining the balance between learning the current task and retaining prior knowledge. Moreover, since the gradient \(g_t\) is also penalized, this approach helps mitigate potential issues such as gradient explosions.

In summary, the weight rollback mechanism effectively constrains the variation between \(\theta_{t+1}\) and \(\theta_0\) at each update step. This not only alleviates the risk of overfitting to the current task but also aids in preserving the knowledge acquired from previous tasks, enhancing the overall performance of the fine-tuning process within the $\text{DualOpt}_{Fine-tune}$ framework.

\subsection{Layer-Wise Penalty for Fine-Tuning}
To further enhance the fine-tuning process, we introduce a layer-wise penalty mechanism. This technique adjusts the penalty applied to the weights based on the layer's depth in the network, ensuring that shallow layers, which capture basic features, receive less stringent penalties compared to deeper layers, which focus on more abstract representations.
Specifically, in deep learning neural networks, each layer can be viewed as a function that processes its input. Given a layer index \(i\), this relationship can be expressed as:
\begin{equation}
    x_{i+1} = f_i(x^*_i),
\end{equation}
where \(f_i\) denotes the \(i\)-th layer. Let \(x^u_i\) represent the input to \(f_i\) in upstream tasks with distribution \(q_i(x^u_i)\), and \(x^d_i\) represent the input in downstream tasks with distribution \(p_i(x^d_i)\). Since \(q_i(x^u_i)\) and \(p_i(x^d_i)\) often differ, we first unfreeze all layers to ensure that \(f_i\) receives sufficient updates to better bridge this gap.

Shallow layers are generally responsible for capturing basic features such as color and texture, while deeper layers focus on more complex features like semantic information \cite{lin2017feature}. This implies that shallow layers are closely tied to data distribution, whereas deep layers align more closely with specific task objectives. A fundamental assumption in transfer learning is that \(q_i(x^u_i)\) shares some similarity with \(p_i(x^d_i)\). Thus, shallow layers tend to exhibit similarities across both pre-training and fine-tuning phases, requiring fewer updates compared to deeper layers.

Based on these insights, we propose a layer-wise penalty decay mechanism for the weight rollback process. This approach gradually reduces the rollback level as the layer depth increases, encouraging shallow layers to extract more general features in downstream tasks while preserving the model’s overall capacity. The penalty factor \(\lambda_i\) for any layer at index \(i\) is calculated using:
\begin{equation}
    \lambda_i = \iota_2 + \left(1 - \frac{i}{n}\right)(\iota_1 - \iota_2),
\end{equation}
where \(n\) is the total number of layers in the pre-trained model, and \(\iota_1\) and \(\iota_2\) denote the maximum and minimum rollback levels, respectively.

Additionally, in different downstream tasks, the objectives may significantly from those of upstream tasks. To accommodate this variability, we propose to adjust the penalty decay rate between layers by applying a power exponent \(\gamma\) to the weight rollback value. This adjustment can be expressed as:
\begin{equation}
    1 - \frac{i}{n} \longrightarrow (1 - \frac{i}{n})^\gamma.
\end{equation}
This dynamic adjustment helps to mitigate the bias arising from a fixed decay rate of the similarities between \(q_i(x^u_i)\) and \(p_i(x^d_i)\) across different layer indices \(i\). Consequently, the penalty decay becomes more adaptable and versatile, catering to requirements dictated by various downstream tasks.

Finally, we provide specific descriptions in Algorithm \ref{algo_sgd} and Algorithm \ref{algo_adam} to combine the weight rollback mechanism and layer-wise penalty in $\text{DualOpt}_{Fine-tune}$ for common optimizers such as SGD and Adam.

	\definecolor{newcolor}{rgb}{0.8,1,1}
	\newcommand{\weightbackcolor}{Thistle}
	\newcommand{\weightback}[1]{\colorbox{\weightbackcolor}{$\displaystyle #1$}}
	\newcommand{\weightbacktext}[1]{\colorbox{\weightbackcolor}{#1}}

	\begin{algorithm}[tb!]
		\caption{$\text{DualOpt}_{Fine-tune}$ for SGD with Momentum}
		\footnotesize
		\label{algo_sgd}
		\begin{algorithmic}[1]
			\STATE{\textbf{input}: \\ $\eta \in \mathbb{R}$: Initial learning rate, $\beta \in [0, 1)$: momentum factor, $\theta_0$: pre-trained weight, $\iota_1,\iota_2 \in [0, 1], \iota_1 \geq \iota_2$: max and min level of weight rollback respectively, $\gamma \in \mathbb{R}$: weight rollback power}
			\STATE{\textbf{initialize}: \\$t \leftarrow 0$: time step, $m_0 \leftarrow 0$: initial moment vector, $d_0 \leftarrow 0$: initial discrepancy value, $\lambda_i \leftarrow f(\lambda, i, n, \iota_1, \iota_2) / \eta$: calculate penalty factor $\lambda_i$ through $\lambda_i = f(\lambda, i, n, \iota_1, \iota_2) = \iota_2+(1-\frac{i}{n})^\gamma(\iota_1-\iota_2)$, then scale it by dividing $\eta$ to eliminate the scale issue.}
			\REPEAT
			\STATE{$t \leftarrow t + 1$}
			\STATE{$\eta_t \leftarrow$$LR_{Scheduler}$$(\eta_{t-1})$ \hfill(Calculate $\eta_t$ at timestep $t$)}
			\STATE{$g_t \leftarrow \nabla_\theta f_t(\theta_{t-1})$ \hfill(Get batch gradient at timestep $t$)} 
			\STATE{$m_t \leftarrow \beta m_{t-1} + (1 - \beta) g_t $\hfill(Compute momentum)}
			\STATE{$\theta_t \leftarrow \theta_{t-1} - \eta_t \lambda_i d_{t-1} - (1-\eta_t \lambda_i)\eta_t m_t$\hfill(Update weight)}
			\STATE{$d_t \leftarrow (1-\eta_t \lambda_i)(d_{t-1}-\eta_t m_t)$\hfill(Update discrepancy)}
			\UNTIL{Stopping condition is met}
			\RETURN{Parameters $\theta_t$}
		\end{algorithmic}
	\end{algorithm}
	
	\begin{algorithm}[tb!]
		\caption{$\text{DualOpt}_{Fine-tune}$ for Adam}
		\footnotesize
		\label{algo_adam}
		\begin{algorithmic}[1]
			\STATE{\textbf{input}: \\ $\eta \in \mathbb{R}$: Initial learning rate, $\beta_1, \beta_2 \in [0, 1)$: Exponential decay rates for the moment estimates, $\epsilon$: bias, $\theta_0$: pre-trained weight, $\iota_1,\iota_2 \in [0, 1], \iota_1 \geq \iota_2$: max and min level of weight rollback respectively, $\gamma \in \mathbb{R}$: weight rollback power}
			\STATE{\textbf{initialize}: \\$t \leftarrow 0$: time step, $m_0 \leftarrow 0$: initial first moment vector, $v_0 \leftarrow 0$: initial second moment vector, $d_0 \leftarrow 0$: initial discrepancy value, $\lambda_i \leftarrow f(\lambda, i, n, \iota_1, \iota_2) / \eta$: calculate penalty factor $\lambda_i$ through $\lambda_i = f(\lambda, i, n, \iota_1, \iota_2) = \iota_2+(1-\frac{i}{n})^\gamma(\iota_1-\iota_2)$, then scale it by dividing $\eta$ to eliminate the scale issue.}
			\REPEAT
			\STATE{$t \leftarrow t + 1$}
			\STATE{$\eta_t \leftarrow$$LR_{Scheduler}$$(\eta_{t-1})$ \hfill(Calculate $\eta_t$ at timestep $t$)}
			\STATE{$g_t \leftarrow \nabla f_t(\theta_{t-1})$\hfill(Get batch gradient at timestep $t$)}
			\STATE{$m_t \leftarrow \beta_1 m_{t-1} + (1 - \beta_1) g_t $\hfill(Update first moment vector)}
			\STATE{$v_t \leftarrow \beta_2 v_{t-1} + (1 - \beta_2) g^2_t $\hfill(Update second moment vector)}
			\STATE{$\hat{m}_t \leftarrow m_t/(1 - \beta_1^t)$}
			\STATE{$\hat{{v}}_t \leftarrow v_t/(1 - \beta_2^t)$}
			\STATE{$\theta_t \leftarrow \theta_{t-1} - \eta_t \lambda_i d_{t-1} - \frac{(1-\eta_t \lambda_i)\eta_t \hat{m}_t}{(\sqrt{\hat{v}_t} + \epsilon)}$\hfill(Update weight)}
			\STATE{$d_t \leftarrow (1-\eta_t \lambda_i)(d_{t-1}-\frac{\eta_t \hat{m}_t}{(\sqrt{\hat{v}_t} + \epsilon)})$\hfill(Update discrepancy)}
			\UNTIL{ stopping criterion is met }
			\RETURN{optimized parameters $\bm{\theta}_t$}
		\end{algorithmic}
	\end{algorithm}

%\section{Experiments}
\section{Experiment Configuration}
\label{sec:experiment}
\subsection{Backbones}
For the training from scratch experiments, we select classic architectures such as ResNet \cite{he2016deep} and Inception \cite{szegedy2015going} as backbones. These architectures are well-established in the field and provide a solid foundation for evaluating the performance of our proposed methods without the benefits of pre-trained weights.

In our fine-tuning experiments, we utilize both CNN-based ConvNeXt \cite{convnext} and Transformer-based Vision Transformers (ViT) \cite{vit} as backbones for the fine-tuning tasks. These models are selected for their representativeness within their respective categories (CNN-based and ViT-based), as they are the most classic architectures with rich pre-training resources. Specifically, we leverage pre-trained weights sourced from several prominent datasets, including ImageNet-1K (MAE) \cite{imagenet-1k}, ImageNet-21K (supervised) \cite{imagenet-21k}, and LAION-2B (CLIP) \cite{laion-2b}. The weights from ImageNet-21K are derived from supervised pre-training, while those from the other datasets utilize self-supervised learning techniques.

\subsection{Benchmarks}
We evaluate our methods across ten widely used visual task datasets, including CIFAR-100 \cite{cifar-100}, SVHN \cite{svhn}, CUB-200-2011 \cite{cub-200}, Stanford Cars \cite{standfordcars}, Places-LT \cite{places-lt}, IP102 \cite{ip102}, OfficeHome \cite{officehome}, PACS \cite{PACS}, COCO2017 \cite{coco} and ADE20K \cite{ade20k}. These datasets encompass a variety of tasks such as general classification, fine-grained classification, long-tailed classification, cross-domain classification, object detection, semantic segmentation, and instance segmentation. A detailed overview of these datasets is presented in Table \ref{tab:dataset}.

\subsection{Baselines}
To facilitate a comprehensive evaluation, we select a range of state-of-the-art methods and traditional approaches as baselines. This includes Full Fine-tuning (Full), Linear Probing (Linear) \cite{linearprob}, L2-SP \cite{L2-SP}, and VPT \cite{VPT}. Following established methods \cite{cnnsgdvitadam}, we employ the SGD optimizer for CNN-based backbones and the Adam optimizer for Transformer-based architectures. 
In contrast, for the training-from-scratch experiments, we focus on evaluating the improvements of our proposed method over the original SGD and Adam optimizers.

\begin{table}
	\centering
    \caption{Details of the used datasets.}
	\begin{tabular}{llll}
		\toprule 
		Dataset         & Images & Categories & Type \\
		\midrule
		CIFAR-100 \cite{cifar-100}  &   60,000 & 100     & General \\
		SVHN \cite{svhn}         & 600,000 & 10     & General \\
		CUB-200-2011 \cite{cub-200}   & 11,788  & 200  & Fine-grained \\
		Stanford Cars \cite{standfordcars}  & 16,185 & 196 &  Fine-grained \\
		Places-LT \cite{places-lt}    & 62,500 & 365  &  Long-tailed \\
		IP102 \cite{ip102}           & 75,222  & 102  &  Long-tailed \\
		OfficeHome \cite{officehome}      & 15,500   & 4 $\times$ 65 &  Cross-domain \\
		PACS \cite{PACS}           & 9,991  & 4 $\times$ 7 &  Cross-domain \\
		COCO2017 \cite{coco}           & 163,957  & 80 &  Detection \\
		ADE20K \cite{ade20k}           & 27,574  & 3688 &  Segmentation \\
		\bottomrule
	\end{tabular}
	\label{tab:dataset}
\end{table}

\subsection{Implementation Details}
% For the training from scratch experiments, the batch size is set to 128, the learning rate is 0.01, and the training duration is 50 epochs. The weight decay is set to 1e-4. The data preprocessing pipeline includes resizing images to \(224 \times 224\), applying random horizontal flips for augmentation.
For the training from scratch experiments, the batch size is set to 128, the learning rate is $10^{-2}$, and the training duration is 50 epochs. The weight decay is set to $10^{-4}$. The data preprocessing pipeline includes resizing images to \(224 \times 224\), applying random horizontal flips for augmentation.
For optimizers, we use either SGD with a momentum of 0.9 or AdamW with the same learning rate and weight decay. The experiments are conducted on an NVIDIA GeForce RTX 4090 GPU, using Python 3.8.10, Ubuntu 20.04, and the PyTorch 2.0.0 framework. 

For the fine-tuning experiments, the input images are resized to \(224 \times 224\) pixels. The batch size is determined by the chosen freezing strategy, with values of 128, 256, and 512 for full unfreezing, parameter isolation, and full freezing methods, respectively.
% For ConvNeXt backbones, we implement the SGD optimizer with a momentum of 0.9. The learning rates vary based on the freezing strategy, set at 1e-2, 2e-2, and 4e-2 for full unfreezing, parameter isolation, and full freezing methods, respectively. For ViT backbones, we utilize the Adam optimizer with momentum parameters (0.9, 0.999), adjusting the learning rates to 1e-4 for full unfreezing, 2e-4 for partial unfreezing, and 4e-4 for full freezing. 
For ConvNeXt backbones, we implement the SGD optimizer with a momentum of 0.9. The learning rates vary based on the freezing strategy, set at $10^{-2}$, $2\times10^{-2}$, and $4\times10^{-2}$ for full unfreezing, parameter isolation, and full freezing methods, respectively. 
For ViT backbones, we utilize the Adam optimizer with momentum parameters (0.9, 0.999), adjusting the learning rates to $10^{-4}$ for full unfreezing, $2\times10^{-4}$ for partial unfreezing, and $4\times10^{-4}$ for full freezing.
Training is conducted for 30 epochs on cross-domain datasets and 50 epochs for other datasets. Experiments are conducted on two NVIDIA A5000 GPUs with 24 GB memory each, running on the Ubuntu 20.04 operating system. We utilize Python 3.8.3 as the programming language and the PyTorch 2.0.0 framework for implementation.

		\begin{table*}
        \caption{Comparison of training from scratch results on various types of classification datasets (general, fine-grained, long-tailed, cross-domain).}
		\centering
		\small
		%toprule midrule bottomrule
		%	\renewcommand{\tabcolsep}{3pt}
		\begin{tabular}{lcccccccc}
			\toprule
			& \multicolumn{2}{c}{General (ID)} & \multicolumn{2}{c}{Fine-Grained (ID)} & \multicolumn{2}{c}{Long-Tailed (OOD)} & \multicolumn{2}{c}{Cross-Domain (OOD)} \\
			\cmidrule(lr){2-3} \cmidrule(lr){4-5} \cmidrule(lr){6-7} \cmidrule(lr){8-9}
			Method & Cifar-100 & SVHN & CUB-200-2011 & StanfordCars & Places-LT & IP102 & OfficeHome & PACS \\
			\midrule
			\multicolumn{3}{l}{ResNet-50}\\
			\cmidrule{1-1}
			Adam & 63.47 & 94.36 & 28.05 & 25.05 & 16.33 & 50.45 & \textbf{52.88} & 74.65\\
			\textbf{$\text{Adam-DualOpt}_{Scratch}$} &	\textbf{64.67} & \textbf{94.96} & \textbf{29.51} & \textbf{26.03} &	\textbf{17.60} &	\textbf{50.63} & 51.06 & \textbf{74.96}\\
            \cmidrule[0.3pt](lr){1-9}
            SGD & 59.93 & 93.83 & 15.59 & 10.57 & 15.38 & 49.77 & \textbf{43.81} & 67.95\\
            \textbf{$\text{SGD-DualOpt}_{Scratch}$} &	\textbf{60.03} & \textbf{94.19} & \textbf{20.19} & \textbf{10.62} &	\textbf{15.60} &	\textbf{49.88} & 42.01 & \textbf{68.78}\\
			\midrule
			\multicolumn{3}{l}{Inception}\\
			\cmidrule{1-1}
			Adam & 54.48 & 94.48 & 26.49 & \textbf{26.45} & 14.49 & 47.87 & 31.15 & 75.48\\
			\textbf{$\text{Adam-DualOpt}_{Scratch}$} &	\textbf{56.15} & \textbf{95.99} & \textbf{26.89} & 24.13 &	\textbf{15.23} &	\textbf{48.09} & \textbf{42.98} & \textbf{76.48}\\
            \cmidrule[0.3pt](lr){1-9}
            SGD & 57.08 & 94.98 & 19.47 & 14.18 & 12.79 & 44.30 & 32.93 & 62.07\\
            \textbf{$\text{SGD-DualOpt}_{Scratch}$} &	\textbf{58.62} & \textbf{95.15} & \textbf{21.78} & \textbf{17.95} &	\textbf{12.90} &	\textbf{44.85} & \textbf{35.75} & \textbf{63.58}\\
			\bottomrule
		\end{tabular}
		\label{tab:class-scratch}
	\end{table*}

\section{Results and Discussions}
\label{sec:results_analysis}
\subsection{Main Results}

\subsubsection{Results on Classification Tasks}
To validate the efficient convergence and robust generalization of DualOpt, we present the results of training from scratch and fine-tuning across 10 popular classification datasets, each with distinct data distributions and characteristics. For the training from scratch experiments, we employ classical backbones and optimizers, specifically ResNet and Inception with Adam and SGD, respectively. During fine-tuning, we utilize ViT-B and ConvNeXt-B as backbones, paired correspondingly with Adam and SGD optimizers. In addition, we also conduct comprehensive comparisons with other state-of-the-art fine-tuning methods.

\paragraph{Training From Scratch}
We first evaluate the performance of DualOpt when training models from scratch on a diverse set of classification datasets, including general (ID), fine-grained (ID), long-tailed (OOD), and cross-domain (OOD) categories. Table~\ref{tab:class-scratch} presents the comparison results using classical architectures, ResNet-50 and Inception, trained with both Adam and SGD optimizers, as well as their enhanced counterparts $\text{DualOpt}_{Scratch}$. 

Across all datasets and models, $\text{DualOpt}_{Scratch}$ consistently outperforms the standard optimizers. For instance, when using ResNet-50 with $\text{Adam-DualOpt}_{Scratch}$, we observe accuracy improvements on CIFAR-100 (from 63.47\% to 64.67\%) and Places-LT (from 16.33\% to 17.60\%). Similar enhancements are noted with $\text{SGD-DualOpt}_{Scratch}$, particularly on fine-grained datasets like CUB-200-2011, where the accuracy increases from 15.59\% to 20.19\%. 

When Inception is used as the backbone, $\text{DualOpt}_{Scratch}$ also enhances performance across almost all classification tasks, achieving better accuracy than standard Adam and SGD. Notably, on the OfficeHome dataset, the proposed method improves the ODD classification accuracy by 11.83\% and  2.82\% compared to Adam and SGD with Inception, respectively.

These results demonstrate that DualOpt effectively enhances convergence and generalization during training from scratch, regardless of the underlying optimizer or network architecture.

\paragraph{Fine-tuning} 
For fine-tuning tasks, we evaluate our proposed methods, $\text{Adam-DualOpt}_{Fine-tune}$ and $\text{SGD-DualOpt}_{Fine-tune}$, against several state-of-the-art fine-tuning baselines pre-trained on ImageNet-21K, including Linear Probing, Full Fine-tuning, L2-SP, VPT, and LoRA. As listed in Table \ref{tab:class-finetune}, our methods achieve new state-of-the-art performance across general, fine-grained, long-tailed, and cross-domain classification tasks.

Under the ViT-B Backbone, $\text{Adam-DualOpt}_{Fine-tune}$ consistently significantly outperforms all baseline methods. Specifically, it achieves accuracies of 92.89\% on CIFAR-100 and 97.35\% on SVHN in general ID datasets, surpassing VPT by 1.40\% and 2.98\%, respectively. In fine-grained ID datasets, it reaches 84.84\% on CUB-200-2011 and 82.02\% on StanfordCars, improving over L2-SP by 3.98\% and 6.47\%.
For long-tailed OOD datasets, Places-LT and IP102 see accuracies of 38.07\% and 75.34\%, outperforming VPT and L2-SP by 1.05\% and 1.59\%, respectively.
% In cross-domain OOD datasets, OfficeHome and PACS achieve 89.05\% and 94.38\%, representing gains of 4.61\% and 6.64\% over the best baselines.
In cross-domain OOD datasets, OfficeHome and PACS achieve 89.05\% and 94.38\%, representing gains of 2.57\% and 6.59\% over the best baselines (VPT and Full Fine-tuning, respectively).
LoRA, as a representative parameter-efficient fine-tuning method, shows competitive performance on several datasets while requiring substantially fewer trainable parameters. However, its overall performance remains consistently below that of DualOpt, indicating that fully leveraging backbone adaptation remains critical for achieving optimal performance across diverse and challenging fine-tuning scenarios.

Since the pre-trained ConvNeXt model is more stable than the ViT structure, there is not much difference between different methods in fine-tuning. However, our $\text{SGD-DualOpt}_{Fine-tune}$ still consistently improves fine-tuning accuracy across all datasets. 

These results demonstrate the robustness and effectiveness of the proposed DualOpt in various tasks.
	\begin{table*}
        \caption{Comparison of fine-tuning results on various types of classification datasets (general, fine-grained, long-tailed, cross-domain).}
		\centering
		\small
		%toprule midrule bottomrule
		%	\renewcommand{\tabcolsep}{3pt}
		\begin{tabular}{lcccccccc}
			\toprule
			& \multicolumn{2}{c}{General (ID)} & \multicolumn{2}{c}{Fine-Grained (ID)} & \multicolumn{2}{c}{Long-Tailed (OOD)} & \multicolumn{2}{c}{Cross-Domain (OOD)} \\
			\cmidrule(lr){2-3} \cmidrule(lr){4-5} \cmidrule(lr){6-7} \cmidrule(lr){8-9}
			Method & Cifar-100 & SVHN & CUB-200-2011 & StanfordCars & Places-LT & IP102 & OfficeHome & PACS \\
			\midrule
			\multicolumn{3}{l}{ViT-B Backbone}\\
			\cmidrule{1-1}
			Linear & 72.50 & 58.79 & 75.01 & 38.03 & 31.95 & 64.93 & 79.96 & 71.88\\
			Full & 87.76 & 97.27 & 81.34 & 75.55 & 31.59 & 74.09 & 84.39 &	87.79\\
			L2-SP & 88.17 &	97.12 &	81.65 &	75.55 &	31.22 &	73.75 &	84.74 &	87.74 \\
			VPT & 91.49 & 94.37 & 81.86	& 58.24 & 37.02	 & 70.41 &	86.48 &	77.44\\
            LoRA & 87.03 & 96.69 & 77.20 & 76.13 & 33.30 & 69.16 & 84.65 & 90.48 \\
			\textbf{$\text{Adam-DualOpt}_{Fine-tune}$} &	\textbf{92.89} & \textbf{97.35} & \textbf{84.84} & \textbf{82.02} &	\textbf{38.07} &	\textbf{75.34} & \textbf{89.05} & \textbf{94.38}\\
			\midrule
			\multicolumn{3}{l}{ConvNeXt-B Backbone}\\
			\cmidrule{1-1}
			Linear & 81.70&	69.21&	87.85&	50.21&	36.41&	70.77&	92.40&	93.46\\
			Full & 92.72&	96.97&	88.59&	88.67&	38.61&	75.01&	91.78&	95.51\\
			L2-SP & 92.84&	97.01&	88.82&	88.83&	38.52&	75.20&	90.61&	95.90 \\
			VPT & 88.71&	81.58&	87.88&	51.58&	36.32&	71.22&	92.31&	93.75\\
            LoRA & 89.37 & 96.86 & 86.16 & 48.82 & 37.11 & 74.10 & 87.65 & 93.70 \\
			\textbf{$\text{SGD-DualOpt}_{Fine-tune}$} &\textbf{92.86}&	\textbf{97.12}&	\textbf{89.47}&	\textbf{88.99}&	\textbf{39.36}&	\textbf{75.44}&	\textbf{92.59}&	\textbf{96.63}\\
			\bottomrule
		\end{tabular}
		\label{tab:class-finetune}
	\end{table*}

\subsubsection{Results on Large-Scale Dataset Classification Tasks}
To evaluate the effectiveness of DualOpt on large-scale datasets, we conducted experiments on ImageNet using two representative transformer-based backbones: ViT-B and Swin-B. As lited in Table~\ref{tab:large-scale}, DualOpt consistently improves Top-1 accuracy compared to the original optimizers. Specifically, for ViT-B, DualOpt achieves an accuracy of 82.58\%, surpassing the baseline by 1.24\%. Similarly, for Swin-B, it improves accuracy from 83.05\% to 83.89\%. These results demonstrate that DualOpt also effectively enhances performance on large-scale datasets, with the potential to provide significant improvements over pre-trained baseline models in deep learning.

 \begin{table}
		\centering
		\small
            \caption{Results of DualOpt for Model Training From Scratch on ImageNet.}
		\begin{tabular}{lccc}
			\toprule
			Method & Model & Dataset& Top-1 Acc \\
			\midrule
			Origin & ViT-B & ImageNet & 81.34\\
			\textbf{DualOpt} & ViT-B & ImageNet & \textbf{82.58}\\
                Origin & Swin-B & ImageNet & 83.05 \\
                \textbf{DualOpt} & Swin-B & ImageNet & \textbf{83.89}\\
			\bottomrule
		\end{tabular}
		\label{tab:large-scale}
\end{table}

 	\begin{table}
		\centering
		\small
            \caption{Results of object detection and instance segmentation using the ConvNeXt-B as backbone.}
		%toprule midrule bottomrule
		%	\renewcommand{\tabcolsep}{3pt}
		\begin{tabular}{lcccc}
			\toprule
			Method & Model & Dataset& $Bbox_m$ & $Segm_m$ \\
			\midrule
			Full & Mask R-CNN & COCO2017 & 40.20 & 36.00 \\
			\textbf{DualOpt} & Mask R-CNN & COCO2017 &\textbf{41.10} & \textbf{36.90} \\
			\bottomrule
		\end{tabular}
		\label{tab:coco2017}
	\end{table}

	\begin{table}
		\centering
		\small
        \caption{Results of semantic segmentation using the ViT-B as backbone.}
		%toprule midrule bottomrule
		\renewcommand{\tabcolsep}{14pt}
		\begin{tabular}{lccc}
			\toprule
			Method & Model & Dataset& $IOU_m$ \\
			\midrule
			Full & UperNet & ADE20K  & 43.65 \\
			\textbf{DualOpt} & UperNet & ADE20K  &\textbf{44.62} \\
			\bottomrule
		\end{tabular}
		\label{tab:ade20k}
	\end{table}
    
\subsubsection{Results on Detection and Segmentation Tasks}
In addition to classification tasks, we evaluate the performance of our DualOpt on object detection, instance segmentation, and semantic segmentation tasks. Due to the complexity of these tasks, the common practice is to import pre-trained weights followed by full fine-tuning. However, our DualOpt approach can easily be applied when integrated with the optimizers. Tables \ref{tab:coco2017} and \ref{tab:ade20k} summarize the results of these experiments.

For object detection and instance segmentation, we utilize the Mask R-CNN model on the COCO2017 dataset. As listed in Table \ref{tab:coco2017}, the proposed DualOpt, using ConvNeXt-B as the backbone, achieves a noticeable improvement over the Full Fine-tuning baseline. Specifically, DualOpt improves the bounding box mean Average Precision (Bbox mAP) from 40.20\% to 41.10\%, and the instance segmentation mAP from 36.00\% to 36.90\%. These results highlight the effectiveness of DualOpt in enhancing object detection and segmentation performance.

Moreover, we employ the UperNet model on the ADE20K dataset for semantic segmentation. As indicated in Table \ref{tab:ade20k}, DualOpt with the ViT-B backbone surpasses the Full Fine-tuning baseline, achieving an improvement in mean Intersection over Union from 43.65\% to 44.62\%. This consistent improvement in segmentation performance demonstrates the adaptability and robustness of DualOpt across different visual tasks and model architectures.

Overall, these results demonstrate that DualOpt not only excels in classification tasks but also provides substantial improvements in object detection, instance segmentation, and semantic segmentation, demonstrating its versatility and effectiveness in more complex scenarios.

\subsection{Analysis and Discussion}
% \begin{table}
% 	\centering
% 	\small
% 	\caption{Comparison of Method efficiency.}
% 	%toprule midrule bottomrule
% 	\renewcommand{\tabcolsep}{3pt}
% 	\begin{tabular}{lccccc}
% 		\toprule 
% 		& & \multicolumn{2}{c}{FLOPs (G)} & \multicolumn{2}{c}{Memory (GB)} \\
% 		\cmidrule(lr){3-4} \cmidrule(lr){5-6}
% 		Method & Params (M) & Train & Inference & Train & Inference \\
% 		\midrule
% 		Linear & 85.69 & \textbf{2156.57} & 2156.54 & \textbf{1.98} & 1.83 \\
% 		VPT& 85.78 & 2482.22 & 2265.26 & 6.63 & 2.12 \\
% 		Full& 85.69 & 6469.62& 2156.54 & 10.45 & 1.83 \\
% 		L2-SP& 85.69 & 6469.77 & 2156.54 & 11.08 & 1.83 \\
% 		DualOpt& 85.69 & 6470.22 & \textbf{2156.54} & 10.90 & \textbf{1.83} \\
% 		\bottomrule
% 	\end{tabular}
% 	\label{tab:efficiency}
% \end{table}
\subsubsection{Efficiency and Complexity Analysis}
\begin{table*}
	\centering
	\small
	\caption{Comparison of method efficiency on Base and Large backbones.}
	\renewcommand{\tabcolsep}{3pt}
	\begin{tabular}{lcccccccccc}
		\toprule
		& \multicolumn{5}{c}{ViT-B Backbone} & \multicolumn{5}{c}{ViT-L Backbone} \\
		\cmidrule(lr){2-6} \cmidrule(lr){7-11}
		Method 
        & Params (M) 
        & FLOPs$_{\text{Tr}}$ 
        & FLOPs$_{\text{Inf}}$ 
        & Mem$_{\text{Tr}}$ 
        & Mem$_{\text{Inf}}$
        & Params (M) 
        & FLOPs$_{\text{Tr}}$ 
        & FLOPs$_{\text{Inf}}$ 
        & Mem$_{\text{Tr}}$ 
        & Mem$_{\text{Inf}}$ \\
		\midrule
		Linear 
        & 85.69 & \textbf{2156.57} & 2156.54 & \textbf{1.98} & 1.83
        & 303.40 & \textbf{7634.73} & 7634.72 & \textbf{2.85} & 2.84 \\

        LoRA
        & 88.63 & 6425.17 & 2159.72 & 9.95 & 1.83 
        & 308.65 & 22844.97 & 7864.99 & 25.25 & 2.84 \\ 
		VPT 
        & 85.78 & 2482.22 & 2265.26 & 6.63 & 2.12
        & 303.42 & 24813.32 & 8271.11 & 17.25 & 2.91 \\
        
		Full 
        & 85.69 & 6469.62 & 2156.54 & 10.45 & 1.83
        & 303.40 & 22884.43 & 7634.72 & 25.95 & 2.84 \\
        
		L2-SP 
        & 85.69 & 6469.77 & 2156.54 & 11.08 & 1.83
        & 303.40 & 22884.43 & 7634.72 & 27.73 & 2.84 \\
        
		DualOpt 
        & \textbf{85.69} & 6470.22 & \textbf{2156.54} & 10.90 & \textbf{1.83}
        & \textbf{303.40} & 22884.43 & \textbf{7634.72} & 27.11 & \textbf{2.84} \\
		\bottomrule
	\end{tabular}
	\label{tab:efficiency-combined}
\end{table*}

In deep learning networks, the computational cost is primarily determined by the forward and backward processes of the network. For DualOpt during training from scratch, the optimization steps and coefficients are adjusted within the optimizer, which does not introduce any additional overhead.

The proposed DualOpt in the fine-tuning version introduces only a minimal computational overhead by adding a review step in the parameter update function of the optimizer during training. This addition maintains theoretical complexity at $O(N)$, ensuring negligible overhead relative to the $O(N^2)$ complexity of forward and backward operations in most networks, and introduces no additional overhead during inference.

As summarized in Table~\ref{tab:efficiency-combined}, we analyze the computational efficiency of different fine-tuning methods on both base and large backbones under a unified experimental setting with a batch size of 128. Compared with the baseline Full Fine-tuning method, DualOpt introduces only 0.04\% extra FLOPs and 4.30\% additional memory usage during training on the base backbone, while incurring 0.00\% extra overhead during inference. A similar efficiency pattern is consistently observed when scaling to the large backbone, demonstrating that DualOpt preserves its efficiency advantages across different model scales.

Although freeze-based methods such as Linear and VPT are more efficient in terms of computational resources, they fall significantly short in performance compared to unfreeze-based methods such as Full Fine-tuning, L2-SP, and DualOpt. For example, while Linear and VPT reduce FLOPs and memory consumption, their fine-tuning performance is substantially lower (as reported in Table~\ref{tab:class-finetune}), highlighting the trade-off between efficiency and performance in fine-tuning tasks.

Overall, DualOpt achieves a favorable balance between efficiency and effectiveness by maintaining competitive computational efficiency while consistently delivering superior fine-tuning performance across different backbone scales.

 \begin{figure*}[htbp]
    \centering
    \includegraphics[width=\textwidth]{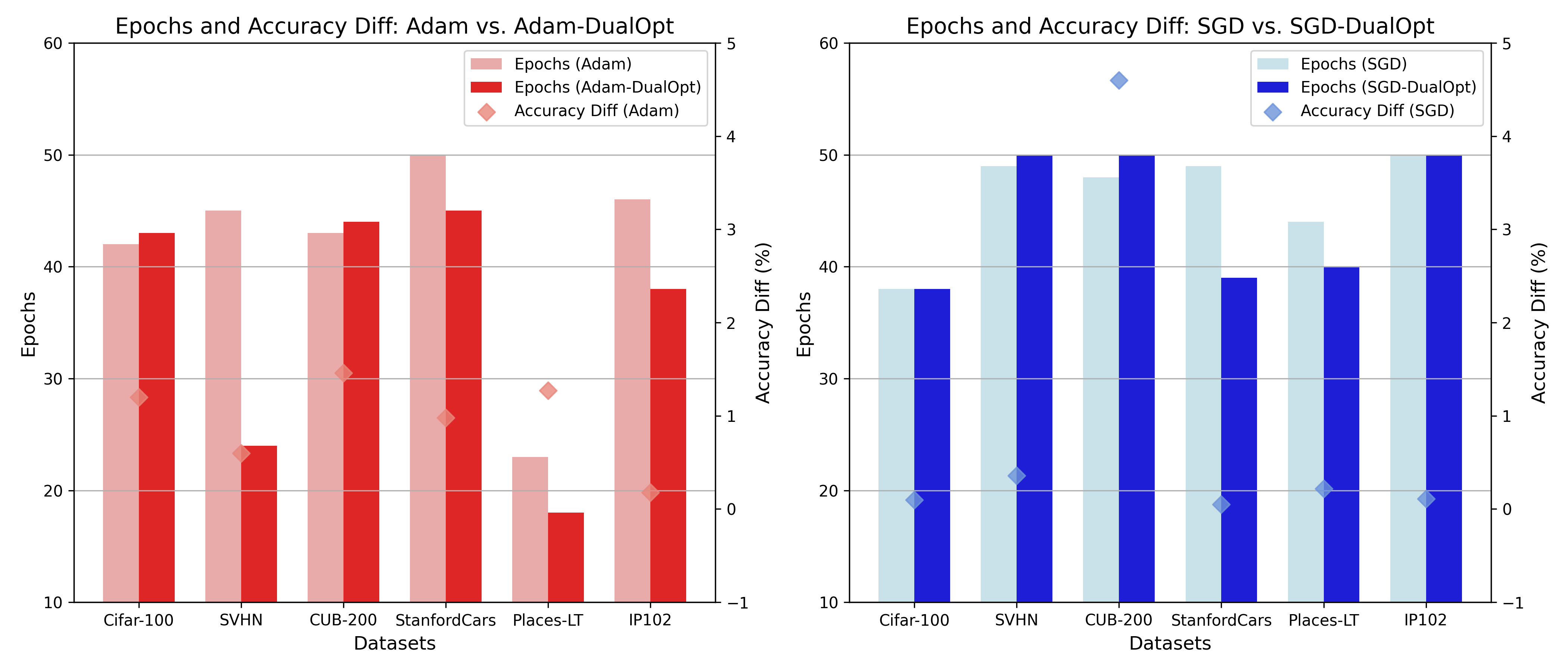}
    \caption{Comparison of epochs and accuracy improvement between standard and DualOpt-enhanced optimizers across different datasets. The left panel shows results for Adam versus Adam-DualOpt, while the right panel displays comparisons for SGD versus SGD-DualOpt. These results highlight DualOpt's ability to significantly reduce training epochs while improving model accuracy.}
    \label{epochs_vs_accuracy_diff_comparison_combined_legend}
\end{figure*}

\subsubsection{Analysis of Performance Across Using Different Large-scale Pre-Trained Models When Fine-tuning}

% To analyze the impact of different pre-trained models on fine-tuning performance, we conduct experiments using three distinct pre-trained ViT-B models: ImageNet-21K (supervised pre-training paradigms), LAION-2B (OpenCLIP, self-supervised pretraining paradigms), and ImageNet-1K (MAE, self-supervised pretraining paradigms). These models are fine-tuned on the challenging PACS dataset to evaluate how various fine-tuning methods adapt to diverse pre-trained weights.
To analyze the compatibility of DualOpt with different pretraining paradigms, we conduct experiments using three distinct ViT-B models pre-trained with large-scale datasets: ImageNet-21K using supervised pretraining, LAION-2B using OpenCLIP with contrastive self-supervised pretraining, and ImageNet-1K using MAE with masked self-supervised pretraining. All models are fine-tuned on the challenging PACS dataset to evaluate how different fine-tuning methods adapt to diverse pre-trained representations.

As listed in Table~\ref{tab:vitopenclipmae}, the proposed DualOpt method outperforms other methods across all pre-trained models. Specifically, DualOpt achieves improvements of 6.59\%, 0.64\%, and 3.47\% over the best-performing baselines when using the Supervised, OpenCLIP, and MAE models, respectively. This consistent superiority indicates that DualOpt adapts more effectively to various pre-trained models, whereas other methods exhibit greater variability in performance. The versatility of DualOpt underscores its robustness and potential for broader application across different pre-training paradigms.

\begin{table}[ht]
    \centering
    \caption{Performance comparison of different fine-tuning methods using various pre-trained models on the PACS dataset.}
    \small
    \renewcommand{\tabcolsep}{5pt}
    \begin{tabular}{lccc}
        \toprule
        Method 
& \makecell{Supervised \\ (ImageNet-21K)}
& \makecell{Self-supervised \\ (OpenCLIP)} 
& \makecell{Self-supervised \\ (MAE)} \\
        % Method & Supervised & OpenCLIP & MAE \\
        \midrule
        Linear & 71.88 & 95.61 & 36.72 \\
        Full   & 87.79 & 47.17 & 84.18 \\
        L2-SP  & 87.74 & 45.56 & 85.79 \\
        VPT    & 77.44 & 97.46 & 50.54 \\
        \textbf{DualOpt} & \textbf{94.38} & \textbf{98.10} & \textbf{89.26} \\
        \bottomrule
    \end{tabular}
    \label{tab:vitopenclipmae}
\end{table}

%\subsubsection{Analysis of Performance with Large-scale Dataset When Training from Scratch}

\subsubsection{Performance and Convergence Analysis When Training from Scratch}
When training from scratch, we theoretically expect improvements in both the efficiency and stability of convergence, along with performance gains. To evaluate the impact of the proposed DualOpt, we compare the number of epochs required for convergence and the performance improvements when integrated with the Adam and SGD optimizers.

As illustrated in Fig \ref{epochs_vs_accuracy_diff_comparison_combined_legend}, when comparing Adam to Adam-DualOpt, our method notably reduces the number of epochs needed for convergence while consistently improving accuracy. This indicates a significant boost in training efficiency and demonstrates the effectiveness of DualOpt in optimizing the training process. Similarly, for SGD and SGD-DualOpt, DualOpt not only shortens convergence times in most cases but also consistently leads to stable performance improvements. Notably, despite iterating through two additional epochs on the CUB-200-2011 dataset, DualOpt achieved an accuracy improvement of up to 4.6\%.

These observations indicate that the proposed DualOpt accelerates convergence while also enhancing model accuracy, confirming its potential as an efficient and effective optimization method for training neural networks from scratch.

\subsubsection{Knowledge Forgetting Test for Fine-tuning}
To evaluate the impact of DualOpt on mitigating knowledge forgetting, we conduct an experiment on the challenging cross-domain PACS dataset, utilizing ViT-B and the Adam optimizer. The dataset is split into two folds: the first fold ($\mathcal{D}_1$) consists of three domains—cartoon, photo, and sketch—while the second fold ($\mathcal{D}_2$) contains data from the art painting domain. During the pre-training stage, the model is trained on $\mathcal{D}_1$ and validated on $\mathcal{D}_2$ for 100 epochs. In the fine-tuning stage, the model is then fine-tuned using $\mathcal{D}_2$ as the training set and $\mathcal{D}_1$ as the validation set for 30 epochs.

\begin{figure}
    \begin{center}
        \centering
        \includegraphics[width=0.9\linewidth]{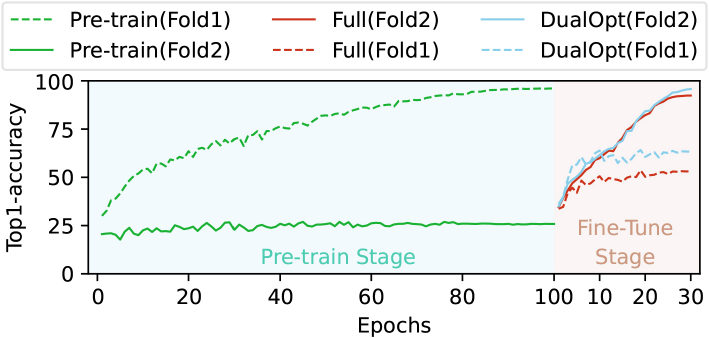}
    \end{center}
    \caption{Knowledge forgetting test on PACS. The first fold is used for pre-training, while the second fold is used for validation during pre-training, and splits are reversed during fine-tuning. DualOpt achieves the best downstream performance while minimizing knowledge forgetting on upstream tasks.}
    \label{fig-review}
\end{figure}

As shown in Fig.~\ref{fig-review}, the Full Fine-tuning baseline improves performance on the downstream task ($\mathcal{D}_2$) but suffers a significant drop in accuracy on upstream domains ($\mathcal{D}_1$), demonstrating severe knowledge forgetting. In contrast, the proposed DualOpt achieves the best performance on $\mathcal{D}_2$ while substantially reducing the accuracy drop on $\mathcal{D}_1$. This result confirms that DualOpt effectively preserves upstream knowledge during fine-tuning and maximizing downstream performance.

%Fig. \ref{fig-review} 展示预训练和微调阶段在Top-1准确率上的变化。使用Full微调方法时，模型在微调后对当前任务的表现提升，但知识遗忘严重。使用OLOR时，模型在微调阶段对当前任务的表现同样提升，同时预训练阶段的知识保留得更好。

% \begin{figure}
%     \begin{center}
%         \centering
%         \includegraphics[width=\linewidth]{figs/train_loss_val_acc.pdf}
%     \end{center}
%     \caption{Comparison of train loss and validation top-1 accuracy on CIFAR-100. DualOpt's performance is compared against standard Adam and SGD, using ViT-B with Adam and ConvNeXt-B with SGD as the backbones.}
%     \label{fig-loss}
% \end{figure}

\begin{figure}
    \begin{center}
        \includegraphics[width=\linewidth]{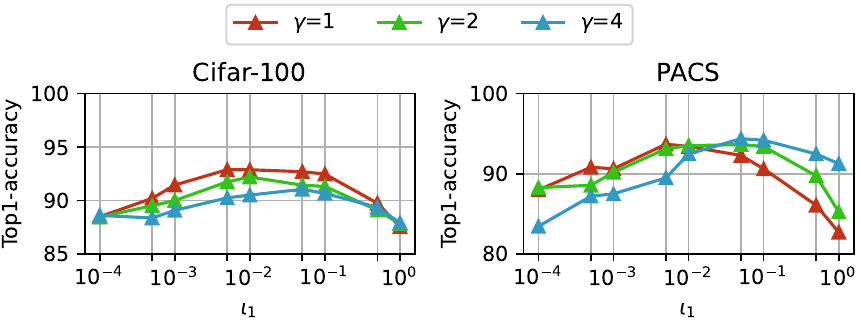}
    \end{center}
    \caption{Hyperparameter exploration experiments in DualOpt fine-tuning mode using ViT-B and Adam on Cifar-100 (left) and PACS (right).}
    \label{fig-hyp}
\end{figure}

% \subsubsection{Compatibility Analysis for Fine-tuning}
% In this section, we further assess the compatibility of DualOpt with different architectures and optimizers in fine-tuning scenarios. As illustrated in Fig. \ref{fig-loss}, integrating weight rollback into various models and optimizers generally improves performance. While the initial convergence speed of the loss may be slower due to the constraints imposed by DualOpt on parameter updates, the final results are highly competitive with the Full Fine-tuning method. This is particularly noticeable with ViT-B optimized using Adam, where DualOpt demonstrates a clear advantage. The validation results reveal that DualOpt not only aids in reducing knowledge forgetting but also leads to a substantial increase in top-1 accuracy over time. This trend becomes especially evident as the model progresses through training.

% This behavior indicates that the weight rollback mechanism in DualOpt plays a pivotal role in preserving important knowledge from pre-training, allowing models to adapt effectively to downstream tasks while minimizing the risk of knowledge forgetting. Moreover, the validation results suggest that DualOpt not only improves model accuracy but also maintains stability across different types of models and optimizers. Overall, these findings affirm that DualOpt is highly compatible with both Transformer-based and CNN-based architectures, and performs well with widely used optimizers.

\begin{figure*}
    \begin{center}
        \includegraphics[width=\linewidth]{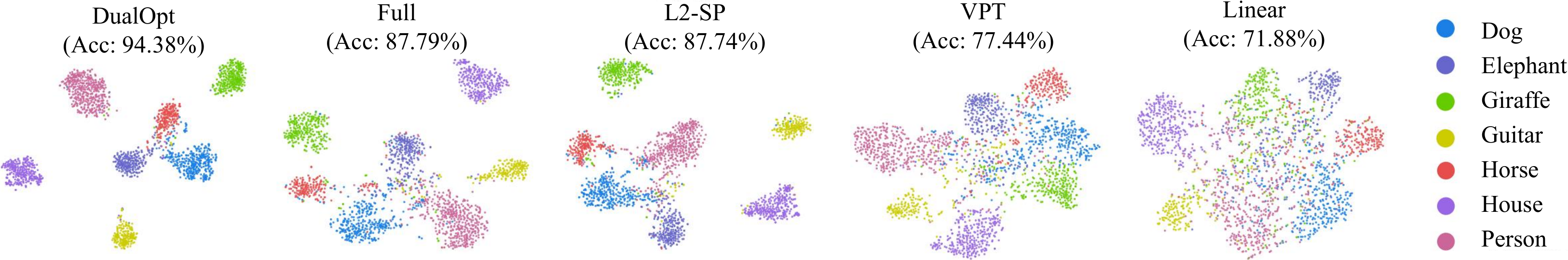}
    \end{center}
    \caption{The t-SNE visualization of the feature distributions on the PACS test set using ViT-B and Adam. The extracted features are color-coded by class, and the accuracy is reported for each method. DualOpt achieves the best feature separation with compact intra-class distances and maximized inter-class distances, resulting in the highest accuracy, while other methods exhibit more overlap between feature clusters and lower accuracy.}
    \label{fig-visual}
\end{figure*}

\subsubsection{Hyper-Parameter Exploration}
To explore the most suitable hyper-parameters of DualOpt for different types of tasks in fine-tuning, we conduct a comprehensive study on both Cifar-100 (ID) and PACS (OOD) datasets. 

Typically, deeper layers in neural networks require significant updates to effectively extract features relevant to the downstream task. Therefore, we simplify the hyperparameter space by setting the minimum rollback level \(\iota_2 = 0\), allowing us to focus on tuning two key hyperparameters: the maximum rollback level \(\iota_1\) and the rollback power \(\gamma\). 
%We explore \(\iota_1\) in the set \(\iota_1 \in \{0.0001, 0.0005, 0.001, 0.005, 0.01, 0.05, 0.1, 0.5, 1\}\), and \(\gamma \in \{1, 2, 4\}\).
The effect of these parameters is explored by setting \(\iota_1\) in the set \(\iota_1 \in \{0.0001, 0.0005, 0.001, 0.005, 0.01, 0.05, 0.1, 0.5, 1\}\), and \(\gamma \in \{1, 2, 4\}\).

Fig. \ref{fig-hyp} shows the results of our hyper-parameter search on Cifar-100 and PACS. We observe that smaller values of $\gamma$ are preferable when the downstream task is similar to the pre-training task, as this allows for more gradual and controlled updates to the deeper layers. Conversely, higher values of $\gamma$ are beneficial when the task distribution changes significantly, allowing the model to adapt more aggressively to new feature distributions. Similarly, for $\iota_1$, larger values are effective when the downstream task's data distribution aligns closely with the upstream task, as they help retain more pre-trained knowledge while fine-tuning.

These findings suggest that the choice of hyper-parameters plays a crucial role in determining the success of fine-tuning, and adjusting $\iota_1$ and $\gamma$ according to task similarity and data distribution can significantly improve performance.

\subsubsection{Feature Quality and Visualization Analysis}
To further evaluate the quality of feature separation and representation learned, we conduct a t-SNE visualization for all fine-tuning methods on the PACS test set based on ViT-B and Adam. 

As shown in Fig. \ref{fig-visual}, DualOpt exhibits superior feature separation compared to the previous methods. The features extracted by DualOpt form well-defined, distinct clusters, with minimal overlap between different classes. This clear separation of feature clusters corresponds to an accuracy of 94.38\%, outperforming all other methods. In contrast, the Full Fine-tuning and L2-SP methods show moderate clustering but still exhibit significant overlap between certain classes, resulting in lower accuracies of 87.79\% and 87.74\%, respectively. VPT and Linear Probing, on the other hand, exhibit poorly separated feature clusters, where the features are highly mixed and result in much lower accuracy.

These results demonstrate that DualOpt improves fine-tuning performance in terms of accuracy and facilitates better feature learning and representation. The well-separated clusters suggest that DualOpt preserves important pre-trained knowledge while adapting to downstream tasks, enabling the model to capture more task-relevant features.

\begin{table}[ht]
    \centering
    \caption{Ablation study of layer-wise weight decay strategies.}
    \small
    \renewcommand{\tabcolsep}{5pt}
    \begin{tabular}{llcc}
        \toprule
        Method & Eq. (\ref{eq:lamda})  & Adam & SGD \\
        \midrule
        Uniform Decay & $\lambda_i = \lambda$ & 63.47 & 59.93 \\
        Proposed Decay  & $\lambda_i = \lambda \frac{i}{n}$ & \textbf{64.67} & \textbf{60.03} \\
        Inverse Proposed Decay    & $\lambda_i = \lambda (1 - \frac{i}{n})$ & 62.61 & 59.67 \\
        \bottomrule
    \end{tabular}
    \label{tab:lamada}
\end{table}

\subsubsection{Impact of Layer-Wise Weight Decay in From-Scratch Training}
In Section \ref{method:FS}, we analyzed the structural characteristics of neural networks during training from scratch. Specifically, shallow layers should use smaller weight decay rates, while deeper layers require larger decay rates. Based on this theory, we proposed a layer-wise weight decay rate $\lambda_i = \lambda \frac{i}{n}$ in Eq. (\ref{eq:lamda}). To evaluate its effectiveness, we conduct an ablation study on the CIFAR-100 dataset using ViT-B with both Adam and SGD optimizers.

The results are summarized in Table~\ref{tab:lamada}. The proposed decay strategy consistently achieves the best performance, with accuracies of 64.67\% and 60.03\% for Adam and SGD, respectively. Notably, when using the inverse strategy, where the weight decay rate is larger for shallow layers and smaller for deeper layers, the performance is even worse than the baseline uniform decay. 
This experiment validates the effectiveness of the proposed layer-wise weight decay mechanism, demonstrating the importance of correctly assigning layer-specific decay rates tailored to the network's depth significantly.

\begin{table}
	\centering
	\small
	\caption{Ablation study of proposed components for DualOpt.}
	%toprule midrule bottomrule
	\renewcommand{\tabcolsep}{4pt}
 	\begin{tabular}{ccccc}
	\toprule 
	Baseline & \makecell{Weight \\ Rollback} & \makecell{Layer-wise\\Penalty} & \makecell{Diversified\\Decay Rate} & Accuracy\\
	\midrule
	\checkmark &&&& 87.79 \\ % Baseline
	& \checkmark &&& 93.41 \\ % Weight Rollback
	& \checkmark & \checkmark && 93.70 \\ % Weight Rollback + Layer-wise Penalty
	& \checkmark & \checkmark & \checkmark & 94.38 \\ % Full DualOpt
	\bottomrule
	\end{tabular}
	\label{tab:ablation_study}
\end{table}

\subsubsection{Effect of Each Component}
In the \textit{Method} Section, we introduce DualOpt, specifically designed to address the distinct needs of both training from scratch and fine-tuning. The fine-tuning component of DualOpt incorporates several key innovations, including the weight rollback mechanism, layer-wise penalty, and diversified decay rate. To evaluate the independent and combined contributions of these components to the overall performance, we conduct an ablation study on the PACS dataset based on Vit-B and Adam. The results are summarized in Table \ref{tab:ablation_study}.

The baseline method, representing standard full fine-tuning without the DualOpt components, achieves an accuracy of 87.79\%. When the weight rollback mechanism is introduced, we observe a substantial improvement in accuracy to 93.41\%, demonstrating the effectiveness of this component in mitigating knowledge forgetting and preserving pre-trained knowledge. The weight rollback mechanism ensures that the model retains essential information from the pre-training stage, allowing it to adapt to new tasks while preventing overfitting to the fine-tuning data.

Adding the layer-wise penalty mechanism to the weight rollback further enhances performance, increasing the accuracy to 93.70\%. This improvement highlights the role of layer-wise penalty in balancing the regularization across different network layers. By applying a more nuanced penalty to different layers based on their depth, the model is better able to adjust to the unique characteristics of downstream tasks. Shallow layers retain more generalized features, while deeper layers are fine-tuned more aggressively, leading to better overall task adaptation.

Finally, incorporating the diversified decay rate results in the best performance, with an accuracy of 94.38\%. The diversified decay rate adjusts the penalty applied to each layer based on the similarity between the pre-training and fine-tuning tasks, further enhancing the model's ability to generalize across different domains. The total improvement of 6.59\% demonstrates that combining all components enables DualOpt to fully leverage pre-trained knowledge while adapting to the specific requirements of downstream tasks.

\section{Conclusion}
\label{sec:conclusion}
In this paper, we proposed DualOpt, a novel optimization framework designed to address the distinct needs of both training from scratch and fine-tuning in neural networks. Unlike traditional optimization methods that treat these two paradigms uniformly, DualOpt decouples the optimization process by introducing specialized strategies for each scenario. For training from scratch, we implemented a real-time layer-wise weight decay mechanism, which dynamically adjusts the decay rates across different layers to enhance convergence speed and generalization. For fine-tuning, we integrated a weight rollback mechanism and a layer-wise penalty decay, ensuring that pre-trained knowledge is preserved while adapting to downstream tasks.
Through extensive experiments on a wide range of visual tasks, including general, fine-grained, long-tailed, cross-domain image classification, object detection, and semantic segmentation, our DualOpt achieves state-of-the-art performance. Extensive validation experiments and ablation analysis demonstrate the effectiveness of the proposed method.
\textbf{Limitations and Future Work}. While DualOpt demonstrates strong empirical performance on a variety of vision tasks, its applicability to other domains, such as large-scale language models or reinforcement learning, has not been fully explored, which remains an interesting direction for future work.
\bibliographystyle{IEEEtran}
\bibliography{ref}

@inproceedings{linearprob,
  title={Split-brain autoencoders: Unsupervised learning by cross-channel prediction},
  author={Zhang, Richard and Isola, Phillip and Efros, Alexei A},
  booktitle={Proceedings of the IEEE Conference on Computer Vision and Pattern Recognition},
  pages={1058--1067},
  year={2017}
}

@inproceedings{openclip,
  title={Learning transferable visual models from natural language supervision},
  author={Radford, Alec and Kim, Jong Wook and Hallacy, Chris and Ramesh, Aditya and Goh, Gabriel and Agarwal, Sandhini and Sastry, Girish and Askell, Amanda and Mishkin, Pamela and Clark, Jack and others},
  booktitle={International {Conference} on {Machine} {Learning}},
  pages={8748--8763},
  year={2021},
  organization={PMLR}
}

@inproceedings{mae,
  title={Masked autoencoders are scalable vision learners},
  author={He, Kaiming and Chen, Xinlei and Xie, Saining and Li, Yanghao and Doll{\'a}r, Piotr and Girshick, Ross},
  booktitle={Proceedings of the IEEE/CVF Conference on Computer Vision and Pattern Recognition},
  pages={16000--16009},
  year={2022}
}

@article{beit,
  title={Beit: Bert pre-training of image transformers},
  author={Bao, Hangbo and Dong, Li and Piao, Songhao and Wei, Furu},
  journal={{ArXiv} {Preprint} arXiv:2106.08254},
  year={2021}
}

@article{laion-2b,
  title={{Laion-5B}: An open large-scale dataset for training next generation image-text models},
  author={Schuhmann, Christoph and Beaumont, Romain and Vencu, Richard and Gordon, Cade and Wightman, Ross and Cherti, Mehdi and Coombes, Theo and Katta, Aarush and Mullis, Clayton and Wortsman, Mitchell and others},
  journal={{ArXiv} {Preprint} arXiv:2210.08402},
  year={2022}
}

@article{laion-400m,
  title={{Laion-400M}: Open dataset of clip-filtered 400 million image-text pairs},
  author={Schuhmann, Christoph and Vencu, Richard and Beaumont, Romain and Kaczmarczyk, Robert and Mullis, Clayton and Katta, Aarush and Coombes, Theo and Jitsev, Jenia and Komatsuzaki, Aran},
  journal={{ArXiv} {Preprint} arXiv:2111.02114},
  year={2021}
}

@article{imagenet-21k,
  title={{ImageNet large} scale visual recognition challenge},
  author={Russakovsky, Olga and Deng, Jia and Su, Hao and Krause, Jonathan and Satheesh, Sanjeev and Ma, Sean and Huang, Zhiheng and Karpathy, Andrej and Khosla, Aditya and Bernstein, Michael and others},
  journal={{International} {Journal} of {Computer} {Vision}},
  volume={115},
  pages={211--252},
  year={2015},
  publisher={Springer}
}

@inproceedings{r_16,
  title={icarl: Incremental classifier and representation learning},
  author={Rebuffi, Sylvestre-Alvise and Kolesnikov, Alexander and Sperl, Georg and Lampert, Christoph H},
  booktitle={Proceedings of the IEEE conference on Computer Vision and Pattern Recognition},
  pages={2001--2010},
  year={2017}
}

@article{r_49,
  title={Experience replay for continual learning},
  author={Rolnick, David and Ahuja, Arun and Schwarz, Jonathan and Lillicrap, Timothy and Wayne, Gregory},
  journal={Advances in Neural Information Processing Systems},
  volume={32},
  year={2019}
}

@article{EWC,
  title={Overcoming catastrophic forgetting in neural networks},
  author={Kirkpatrick, James and Pascanu, Razvan and Rabinowitz, Neil and Veness, Joel and Desjardins, Guillaume and Rusu, Andrei A and Milan, Kieran and Quan, John and Ramalho, Tiago and Grabska-Barwinska, Agnieszka and others},
  journal={Proceedings of the National Academy of Sciences},
  volume={114},
  number={13},
  pages={3521--3526},
  year={2017},
  publisher={National Acad Sciences}
}

@inproceedings{L2-SP,
  title={Explicit inductive bias for transfer learning with convolutional networks},
  author={Xuhong, LI and Grandvalet, Yves and Davoine, Franck},
  booktitle={International Conference on Machine Learning},
  pages={2825--2834},
  year={2018},
  organization={PMLR}
}

@inproceedings{AdamW_2,
  title={Weight Prediction Boosts the Convergence of AdamW},
  author={Guan, Lei},
  booktitle={Pacific-Asia Conference on Knowledge Discovery and Data Mining},
  pages={329--340},
  year={2023},
  organization={Springer}
}

@article{AdamW,
  title={Decoupled weight decay regularization},
  author={Loshchilov, Ilya and Hutter, Frank},
  journal={ArXiv Preprint arXiv:1711.05101},
  year={2017}
}

@inproceedings{VPT_cvpr,
  title={Visual prompt tuning for generative transfer learning},
  author={Sohn, Kihyuk and Chang, Huiwen and Lezama, Jos{\'e} and Polania, Luisa and Zhang, Han and Hao, Yuan and Essa, Irfan and Jiang, Lu},
  booktitle={Proceedings of the IEEE/CVF Conference on Computer Vision and Pattern Recognition},
  pages={19840--19851},
  year={2023}
}

@inproceedings{VPT,
  title={Visual prompt tuning},
  author={Jia, Menglin and Tang, Luming and Chen, Bor-Chun and Cardie, Claire and Belongie, Serge and Hariharan, Bharath and Lim, Ser-Nam},
  booktitle={European Conference on Computer Vision},
  pages={709--727},
  year={2022},
  organization={Springer}
}

@article{de2021continual,
  title={A continual learning survey: Defying forgetting in classification tasks},
  author={De Lange, Matthias and Aljundi, Rahaf and Masana, Marc and Parisot, Sarah and Jia, Xu and Leonardis, Ale{\v{s}} and Slabaugh, Gregory and Tuytelaars, Tinne},
  journal={IEEE Transactions on Pattern Analysis and Machine Intelligence},
  volume={44},
  number={7},
  pages={3366--3385},
  year={2021},
  publisher={IEEE}
}

@article{vit,
  title={An image is worth 16x16 words: Transformers for image recognition at scale},
  author={Dosovitskiy, Alexey and Beyer, Lucas and Kolesnikov, Alexander and Weissenborn, Dirk and Zhai, Xiaohua and Unterthiner, Thomas and Dehghani, Mostafa and Minderer, Matthias and Heigold, Georg and Gelly, Sylvain and others},
  journal={{ArXiv} {Preprint} arXiv:2010.11929},
  year={2020}
}

@inproceedings{convnext,
  title={A convnet for the 2020s},
  author={Liu, Zhuang and Mao, Hanzi and Wu, Chao-Yuan and Feichtenhofer, Christoph and Darrell, Trevor and Xie, Saining},
  booktitle={Proceedings of the IEEE/CVF Conference on Computer Vision and Pattern Recognition},
  pages={11976--11986},
  year={2022}
}

@inproceedings{replay2020,
  title={Generative feature replay for class-incremental learning},
  author={Liu, Xialei and Wu, Chenshen and Menta, Mikel and Herranz, Luis and Raducanu, Bogdan and Bagdanov, Andrew D and Jui, Shangling and de Weijer, Joost van},
  booktitle={Proceedings of the IEEE/CVF Conference on Computer Vision and Pattern Recognition Workshops},
  pages={226--227},
  year={2020}
}

@inproceedings{replay2022,
  title={Practical recommendations for replay-based continual learning methods},
  author={Merlin, Gabriele and Lomonaco, Vincenzo and Cossu, Andrea and Carta, Antonio and Bacciu, Davide},
  booktitle={International Conference on Image Analysis and Processing},
  pages={548--559},
  year={2022},
  organization={Springer}
}

@article{sgd,
  title={Improving generalization performance by switching from adam to sgd},
  author={Keskar, Nitish Shirish and Socher, Richard},
  journal={ArXiv Preprint arXiv:1712.07628},
  year={2017}
}

@article{cifar-100,
  title={Learning multiple layers of features from tiny images},
  author={Krizhevsky, Alex and Hinton, Geoffrey and others},
  year={2009},
  publisher={Toronto, ON, Canada}
}

@article{svhn,
  title={Reading digits in natural images with unsupervised feature learning},
  author={Netzer, Yuval and Wang, Tao and Coates, Adam and Bissacco, Alessandro and Wu, Bo and Ng, Andrew Y},
  year={2011}
}

@article{cub-200,
  title={The caltech-ucsd birds-200-2011 dataset},
  author={Wah, Catherine and Branson, Steve and Welinder, Peter and Perona, Pietro and Belongie, Serge},
  year={2011},
  publisher={California Institute of Technology}
}

@inproceedings{standfordcars,
  title={{3D} object representations for fine-grained categorization},
  author={Krause, Jonathan and Stark, Michael and Deng, Jia and Fei-Fei, Li},
  booktitle={Proceedings of the IEEE International Conference on Computer Vision Workshops},
  pages={554--561},
  year={2013}
}

@article{places-lt,
  title={Learning deep features for scene recognition using places database},
  author={Zhou, Bolei and Lapedriza, Agata and Xiao, Jianxiong and Torralba, Antonio and Oliva, Aude},
  journal={Advances in Neural Information Processing Systems},
  volume={27},
  year={2014}
}

@article{ip102,
  title={The sun attribute database: Beyond categories for deeper scene understanding},
  author={Patterson, Genevieve and Xu, Chen and Su, Hang and Hays, James},
  journal={International Journal of Computer Vision},
  volume={108},
  pages={59--81},
  year={2014},
  publisher={Springer}
}

@inproceedings{officehome,
  title={Deep hashing network for unsupervised domain adaptation},
  author={Venkateswara, Hemanth and Eusebio, Jose and Chakraborty, Shayok and Panchanathan, Sethuraman},
  booktitle={Proceedings of the IEEE Conference on Computer Vision and Pattern Recognition},
  pages={5018--5027},
  year={2017}
}

@inproceedings{PACS,
  title={Deeper, broader and artier domain generalization},
  author={Li, Da and Yang, Yongxin and Song, Yi-Zhe and Hospedales, Timothy M},
  booktitle={Proceedings of the IEEE International Conference on Computer Vision},
  pages={5542--5550},
  year={2017}
}

@inproceedings{imagenet-1k,
  title={{ImageNet}: A large-scale hierarchical image database},
  author={Deng, Jia and Dong, Wei and Socher, Richard and Li, Li-Jia and Li, Kai and Fei-Fei, Li},
  booktitle={2009 IEEE {Conference} on {Computer} {Vision} and {Pattern} {Recognition}},
  pages={248--255},
  year={2009},
  organization={Ieee}
}

@inproceedings{cnnsgdvitadam,
  title={End-to-end object detection with transformers},
  author={Carion, Nicolas and Massa, Francisco and Synnaeve, Gabriel and Usunier, Nicolas and Kirillov, Alexander and Zagoruyko, Sergey},
  booktitle={European Conference on Computer Vision},
  pages={213--229},
  year={2020},
  organization={Springer}
}

@inproceedings{lin2017feature,
  title={Feature pyramid networks for object detection},
  author={Lin, Tsung-Yi and Doll{\'a}r, Piotr and Girshick, Ross and He, Kaiming and Hariharan, Bharath and Belongie, Serge},
  booktitle={Proceedings of the IEEE Conference on Computer Vision and Pattern Recognition},
  pages={2117--2125},
  year={2017}
}

@article{li2023embracing,
  title={Embracing large natural data: Enhancing medical image analysis via cross-domain fine-tuning},
  author={Li, Qiankun and Huang, Xiaolong and Fang, Bo and Chen, Huabao and Ding, Siyuan and Liu, Xu},
  journal={IEEE Journal of Biomedical and Health Informatics},
  year={2023},
  publisher={IEEE}
}

@inproceedings{huang2024one,
  title={One Step Learning, One Step Review},
  author={Huang, Xiaolong and Li, Qiankun and Li, Xueran and Gao, Xuesong},
  booktitle={Proceedings of the AAAI Conference on Artificial Intelligence},
  volume={38},
  number={11},
  pages={12644--12652},
  year={2024}
}

@inproceedings{li2023data,
  title={Data-efficient masked video modeling for self-supervised action recognition},
  author={Li, Qiankun and Huang, Xiaolong and Wan, Zhifan and Hu, Lanqing and Wu, Shuzhe and Zhang, Jie and Shan, Shiguang and Wang, Zengfu},
  booktitle={Proceedings of the 31st ACM International Conference on Multimedia},
  pages={2723--2733},
  year={2023}
}

@inproceedings{li2024advancing,
title={Advancing Micro-Action Recognition with Multi-Auxiliary Heads and Hybrid Loss Optimization},
author={Li, Qiankun and Huang, Xiaolong and Chen, Huabao and He, Feng and Chen, Qiupu and Wang, Zengfu},
booktitle={Proceedings of the 32nd ACM International Conference on Multimedia},
year={2024}
}

@ARTICLE{transfer-pami1,
  author={Mensink, Thomas and Uijlings, Jasper and Kuznetsova, Alina and Gygli, Michael and Ferrari, Vittorio},
  journal={IEEE Transactions on Pattern Analysis and Machine Intelligence}, 
  title={Factors of Influence for Transfer Learning Across Diverse Appearance Domains and Task Types}, 
  year={2022},
  volume={44},
  number={12},
  pages={9298-9314},}

@ARTICLE{transfer-pami2,
  author={Wu, Zuxuan and Weng, Zejia and Peng, Wujian and Yang, Xitong and Li, Ang and Davis, Larry S. and Jiang, Yu-Gang},
  journal={IEEE Transactions on Pattern Analysis and Machine Intelligence}, 
  title={Building an Open-Vocabulary Video CLIP Model With Better Architectures, Optimization and Data}, 
  year={2024},
  volume={46},
  number={7},
  pages={4747-4762},}

@article{strach-pami1,
  title={Recognizing object by components with human prior knowledge enhances adversarial robustness of deep neural networks},
  author={Li, Xiao and Wang, Ziqi and Zhang, Bo and Sun, Fuchun and Hu, Xiaolin},
  journal={IEEE Transactions on Pattern Analysis and Machine Intelligence},
  volume={45},
  number={7},
  pages={8861--8873},
  year={2023},
  publisher={IEEE}
}

@article{strach-pami2,
  title={Medical image segmentation review: The success of u-net},
  author={Azad, Reza and Aghdam, Ehsan Khodapanah and Rauland, Amelie and Jia, Yiwei and Avval, Atlas Haddadi and Bozorgpour, Afshin and Karimijafarbigloo, Sanaz and Cohen, Joseph Paul and Adeli, Ehsan and Merhof, Dorit},
  journal={IEEE Transactions on Pattern Analysis and Machine Intelligence},
  year={2024},
  publisher={IEEE}
}

@misc{kosson2023rotational,
      title={Rotational Equilibrium: How Weight Decay Balances Learning Across Neural Networks}, 
      author={Atli Kosson and Bettina Messmer and Martin Jaggi},
      year={2024},
      eprint={2305.17212},
      archivePrefix={arXiv},
      primaryClass={cs.LG},
      url={https://arxiv.org/abs/2305.17212}, 
}

@inproceedings{xie2021positive,
  title={Positive-negative momentum: Manipulating stochastic gradient noise to improve generalization},
  author={Xie, Zeke and Yuan, Li and Zhu, Zhanxing and Sugiyama, Masashi},
  booktitle={Proc. International Conference on Machine Learning},
  pages={11448--11458},
  year={2021}
}

@article{ghiasi2024improving,
  title={Improving robustness with adaptive weight decay},
  author={Ghiasi, Mohammad Amin and Shafahi, Ali and Ardekani, Reza},
  journal={Proc. Neural Information Processing System},
  volume={36},
  year={2024}
}

@article{tian2024rethinking,
  title={Rethinking weight decay for robust fine-tuning of foundation models},
  author={Tian, Junjiao and Huang, Chengyue and Kira, Zsolt},
  journal={Advances in Neural Information Processing Systems},
  volume={37},
  pages={22418--22440},
  year={2024}
}

@inproceedings{hanson1988comparing,
  title={Comparing biases for minimal network construction with back-propagation},
  author={Hanson, Stephen and Pratt, Lorien},
  journal={Proc. Neural Information Processing Systems},
  volume={1},
  year={1988}
}

@misc{ergen2023globally,
      title={Globally Optimal Training of Neural Networks with Threshold Activation Functions}, 
      author={Tolga Ergen and Halil Ibrahim Gulluk and Jonathan Lacotte and Mert Pilanci},
      year={2023},
      eprint={2303.03382},
      archivePrefix={arXiv},
      primaryClass={cs.LG},
      url={https://arxiv.org/abs/2303.03382}, 
}

@misc{stock2019equi,
      title={Equi-normalization of Neural Networks}, 
      author={Pierre Stock and Benjamin Graham and Rémi Gribonval and Hervé Jégou},
      year={2019},
      eprint={1902.10416},
      archivePrefix={arXiv},
      primaryClass={cs.CV},
      url={https://arxiv.org/abs/1902.10416}, 
}

@article{adam,
  title={Adam: A method for stochastic optimization},
  author={Kingma, Diederik P},
  journal={arXiv preprint arXiv:1412.6980},
  year={2014}
}

@article{mom,
  title={Learning representations by back-propagating errors},
  author={Rumelhart, David E and Hinton, Geoffrey E and Williams, Ronald J},
  journal={nature},
  volume={323},
  number={6088},
  pages={533--536},
  year={1986},
  publisher={Nature Publishing Group UK London}
}

@article{duchi2011adaptive,
  title={Adaptive subgradient methods for online learning and stochastic optimization.},
  author={Duchi, John and Hazan, Elad and Singer, Yoram},
  journal={Journal of Machine Learning Research},
  volume={12},
  number={7},
  year={2011}
}

@article{krogh1991simple,
  title={A simple weight decay can improve generalization},
  author={Krogh, Anders and Hertz, John},
  journal={Advances in Neural Information Processing Systems},
  volume={4},
  year={1991}
}

@article{ade20k,
  title={Semantic understanding of scenes through the ade20k dataset},
  author={Zhou, Bolei and Zhao, Hang and Puig, Xavier and Xiao, Tete and Fidler, Sanja and Barriuso, Adela and Torralba, Antonio},
  journal={International Journal of Computer Vision},
  volume={127},
  pages={302--321},
  year={2019},
  publisher={Springer}
}

@inproceedings{coco,
  title={Microsoft coco: Common objects in context},
  author={Lin, Tsung-Yi and Maire, Michael and Belongie, Serge and Hays, James and Perona, Pietro and Ramanan, Deva and Doll{\'a}r, Piotr and Zitnick, C Lawrence},
  booktitle={Computer Vision--ECCV 2014: 13th European Conference, Zurich, Switzerland, September 6-12, 2014, Proceedings, Part V 13},
  pages={740--755},
  year={2014},
  organization={Springer}
}

@inproceedings{szegedy2015going,
  title={Going deeper with convolutions},
  author={Szegedy, Christian and Liu, Wei and Jia, Yangqing and Sermanet, Pierre and Reed, Scott and Anguelov, Dragomir and Erhan, Dumitru and Vanhoucke, Vincent and Rabinovich, Andrew},
  booktitle={Proceedings of the IEEE Conference on Computer Vision and Pattern Recognition},
  pages={1--9},
  year={2015}
}

@inproceedings{he2016deep,
  title={Deep residual learning for image recognition},
  author={He, Kaiming and Zhang, Xiangyu and Ren, Shaoqing and Sun, Jian},
  booktitle={Proceedings of the IEEE Conference on Computer Vision and Pattern Recognition},
  pages={770--778},
  year={2016}
}

@article{robbins1951stochastic,
  title={A stochastic approximation method},
  author={Robbins, Herbert and Monro, Sutton},
  journal={The Annals of Mathematical Statistics},
  pages={400--407},
  year={1951},
  publisher={JSTOR}
}

@inproceedings{sutskever2013importance,
  title={On the importance of initialization and momentum in deep learning},
  author={Sutskever, Ilya and Martens, James and Dahl, George and Hinton, Geoffrey},
  booktitle={International Conference on Machine Learning},
  pages={1139--1147},
  year={2013},
  organization={PMLR}
}

@article{liu2019variance,
  title={On the variance of the adaptive learning rate and beyond},
  author={Liu, Liyuan and Jiang, Haoming and He, Pengcheng and Chen, Weizhu and Liu, Xiaodong and Gao, Jianfeng and Han, Jiawei},
  journal={arXiv preprint arXiv:1908.03265},
  year={2019}
}

@article{luo2019adaptive,
  title={Adaptive gradient methods with dynamic bound of learning rate},
  author={Luo, Liangchen and Xiong, Yuanhao and Liu, Yan and Sun, Xu},
  journal={arXiv preprint arXiv:1902.09843},
  year={2019}
}

@article{devlin2018bert,
  title={Bert: Pre-training of deep bidirectional transformers for language understanding},
  author={Devlin, Jacob},
  journal={arXiv preprint arXiv:1810.04805},
  year={2018}
}

@article{brown2020language,
  title={Language models are few-shot learners},
  author={Brown, Tom B},
  journal={arXiv preprint arXiv:2005.14165},
  year={2020}
}

@article{bommasani2021opportunities,
  title={On the opportunities and risks of foundation models},
  author={Bommasani, Rishi and Hudson, Drew A and Adeli, Ehsan and Altman, Russ and Arora, Simran and von Arx, Sydney and Bernstein, Michael S and Bohg, Jeannette and Bosselut, Antoine and Brunskill, Emma and others},
  journal={arXiv preprint arXiv:2108.07258},
  year={2021}
}

@inproceedings{deng2009imagenet,
  title={Imagenet: A large-scale hierarchical image database},
  author={Deng, Jia and Dong, Wei and Socher, Richard and Li, Li-Jia and Li, Kai and Fei-Fei, Li},
  booktitle={2009 IEEE conference on computer vision and pattern recognition},
  pages={248--255},
  year={2009},
  organization={Ieee}
}

@article{kuznetsova2020open,
  title={The open images dataset v4: Unified image classification, object detection, and visual relationship detection at scale},
  author={Kuznetsova, Alina and Rom, Hassan and Alldrin, Neil and Uijlings, Jasper and Krasin, Ivan and Pont-Tuset, Jordi and Kamali, Shahab and Popov, Stefan and Malloci, Matteo and Kolesnikov, Alexander and others},
  journal={International journal of computer vision},
  volume={128},
  number={7},
  pages={1956--1981},
  year={2020},
  publisher={Springer}
}

@article{T5,
  title={Exploring the limits of transfer learning with a unified text-to-text transformer},
  author={Raffel, Colin and Shazeer, Noam and Roberts, Adam and Lee, Katherine and Narang, Sharan and Matena, Michael and Zhou, Yanqi and Li, Wei and Liu, Peter J},
  journal={Journal of machine learning research},
  volume={21},
  number={140},
  pages={1--67},
  year={2020}
}

@article{dodge2021documenting,
  title={Documenting large webtext corpora: A case study on the colossal clean crawled corpus},
  author={Dodge, Jesse and Sap, Maarten and Marasovi{\'c}, Ana and Agnew, William and Ilharco, Gabriel and Groeneveld, Dirk and Mitchell, Margaret and Gardner, Matt},
  journal={arXiv preprint arXiv:2104.08758},
  year={2021}
}

@article{zhu2015aligning,
  title={Aligning Books and Movies: Towards Story-like Visual Explanations by Watching Movies and Reading Books},
  author={Zhu, Yukun},
  journal={arXiv preprint arXiv:1506.06724},
  year={2015}
}

@article{hsu2018re,
  title={Re-evaluating continual learning scenarios: A categorization and case for strong baselines},
  author={Hsu, Y},
  journal={arXiv preprint arXiv:1810.12488},
  year={2018}
}

@article{wu2018incremental,
  title={Incremental classifier learning with generative adversarial networks},
  author={Wu, Yue and Chen, Yinpeng and Wang, Lijuan and Ye, Yuancheng and Liu, Zicheng and Guo, Yandong and Zhang, Zhengyou and Fu, Yun},
  journal={arXiv preprint arXiv:1802.00853},
  year={2018}
}

@article{zou2025rhythmformer,
  title={RhythmFormer: Extracting patterned rPPG signals based on periodic sparse attention},
  author={Zou, Bochao and Guo, Zizheng and Chen, Jiansheng and Zhuo, Junbao and Huang, Weiran and Ma, Huimin},
  journal={Pattern Recognition},
  volume={164},
  pages={111511},
  year={2025},
  publisher={Elsevier}
}

@inproceedings{zou2025rhythmmamba,
  title={RhythmMamba: Fast, Lightweight, and Accurate Remote Physiological Measurement},
  author={Zou, Bochao and Guo, Zizheng and Hu, Xiaocheng and Ma, Huimin},
  booktitle={Proceedings of the AAAI Conference on Artificial Intelligence},
  volume={39},
  number={10},
  pages={11077--11085},
  year={2025}
}

@inproceedings{zenke2017continual,
  title={Continual learning through synaptic intelligence},
  author={Zenke, Friedemann and Poole, Ben and Ganguli, Surya},
  booktitle={International conference on machine learning},
  pages={3987--3995},
  year={2017},
  organization={PMLR}
}

@article{li2017learning,
  title={Learning without forgetting},
  author={Li, Zhizhong and Hoiem, Derek},
  journal={IEEE transactions on pattern analysis and machine intelligence},
  volume={40},
  number={12},
  pages={2935--2947},
  year={2017},
  publisher={IEEE}
}

@inproceedings{rannen2017encoder,
  title={Encoder based lifelong learning},
  author={Rannen, Amal and Aljundi, Rahaf and Blaschko, Matthew B and Tuytelaars, Tinne},
  booktitle={Proceedings of the IEEE International Conference on Computer Vision},
  pages={1320--1328},
  year={2017}
}

@inproceedings{houlsby2019parameter,
  title={Parameter-efficient transfer learning for NLP},
  author={Houlsby, Neil and Giurgiu, Andrei and Jastrzebski, Stanislaw and Morrone, Bruna and De Laroussilhe, Quentin and Gesmundo, Andrea and Attariyan, Mona and Gelly, Sylvain},
  booktitle={International Conference on Machine Learning},
  pages={2790--2799},
  year={2019},
  organization={PMLR}
}

@article{hu2021lora,
  title={Lora: Low-rank adaptation of large language models},
  author={Hu, Edward J and Shen, Yelong and Wallis, Phillip and Allen-Zhu, Zeyuan and Li, Yuanzhi and Wang, Shean and Wang, Lu and Chen, Weizhu},
  journal={arXiv preprint arXiv:2106.09685},
  year={2021}
}

@article{pfeiffer2020adapterfusion,
  title={Adapterfusion: Non-destructive task composition for transfer learning},
  author={Pfeiffer, Jonas and Kamath, Aishwarya and R{\"u}ckl{\'e}, Andreas and Cho, Kyunghyun and Gurevych, Iryna},
  journal={arXiv preprint arXiv:2005.00247},
  year={2020}
}

@article{lecun2015deep,
  title={Deep learning},
  author={LeCun, Yann and Bengio, Yoshua and Hinton, Geoffrey},
  journal={nature},
  volume={521},
  number={7553},
  pages={436--444},
  year={2015},
  publisher={Nature Publishing Group UK London}
}

\clearpage
\begin{IEEEbiography}[{\includegraphics[width=1in,height=1.25in,clip,keepaspectratio]{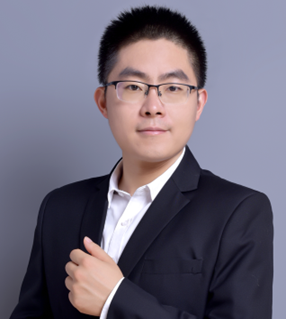}}]{Xin Ning} (Senior Member, IEEE) received the B.S. degree in software engineering from Xinjiang University in 2012, and the Ph.D. degree in computer vision from university of Chinese Academy of Sciences, in 2017. He is currently a Professor with the AnnLab, Institute of Semiconductors, Chinese Academy of Sciences. His current research interests include neural networks, intelligent systems and computer vision. He is working as Guest Editor, Associate Editor in SCI and various other reputed journals (IEEE, Elsevier, Springer \& Wiley). His Google Scholar total citation reached 4500+, with an H-index of 40. He has been featured in the list of top 2\% scientist/researcher database in the world in 2022 and 2023. He has published by first or corresponding author more than 80 papers in journals and refereed conferences. He served as senior member of IEEE/CCF/CAAI/CSIG. He serves as Aare Editor in the Information Fusion and Associate Editor in the IEEE TCSVT, Pattern Recognition, Applied Soft Computing, Engineering Applications of Artificial Intelligence and Signal Processing journal. And he also serves as the Young Associate Editor of CAAI Transactions on Intelligent Systems, the Guest Editor of IEEE TCE, Image and Vision Computing et al. He was the Website Chair of the IEEE HPBD\&IS 2020, and the Publication Chair of the IEEE HPBD\&IS 2021, HDIS 2022 and HDIS 2023. \end{IEEEbiography}

\begin{IEEEbiography}[{\includegraphics[width=1in,height=1.25in,clip,keepaspectratio]{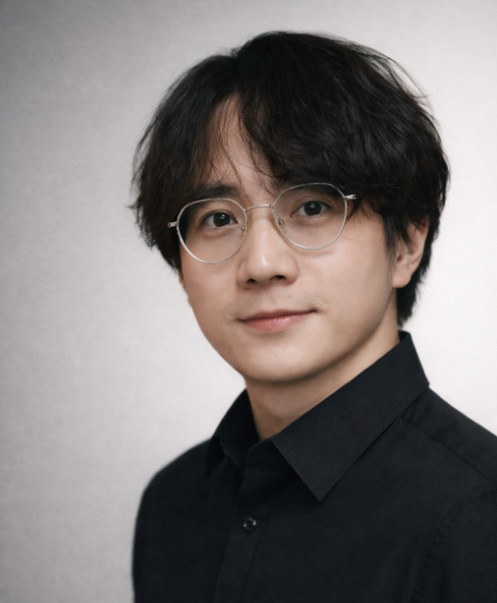}}]{Qiankun Li} (Member, IEEE)
received the Ph.D. degree in Pattern Recognition and Intelligent Systems from the University of Science and Technology of China (USTC), in 2025, where he was with the National Engineering Research Center of Speech and Language Information Processing (NERC-SLIP), Department of Automation. 
He is currently a Research Fellow at Imperial Global Singapore (IGS), an affiliation of Imperial College London (IC). Prior to this, he was a Research Fellow at the College of Computing and Data Science (CCDS) at Nanyang Technological University (NTU). 
His research interests include human-centered AI, particularly AI for healthcare (AI4Health), multimodal large model reasoning and hallucination mitigation, and cross-domain model generalization. He has published several papers in prestigious journals and conference proceedings such as ICML, NeurIPS, CVPR, AAAI, ACM MM, IEEE T-PAMI, IEEE T-IP, IEEE T-FS, IEEE T-ITS, PR, and INFFUS, etc.
\end{IEEEbiography}

\begin{IEEEbiography}[{\includegraphics[width=1in,height=1.25in,clip,keepaspectratio]{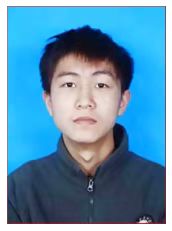}}]{Xiaolong Huang} is currently pursuing a Master's degree at the Gina Cody School of Engineering and Computer Science, Concordia University, Montreal, QC, Canada. He is also affiliated with the Mila-Quebec AI Institute, Montreal, Canada. His research interests generally lie in machine learning and computer vision. \end{IEEEbiography}

\begin{IEEEbiography}[{\includegraphics[width=1in,height=1.25in,clip,keepaspectratio]{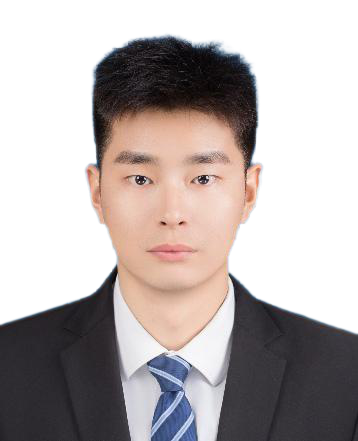}}]{Qiupu Chen} received the Ph.D. degree in Pattern Recognition and Intelligent Systems from the University of Science and Technology of China (USTC), in 2025. 
He is currently an Assistant Professor with the School of Artificial Intelligence, Henan Univeristy.
His research interests include human-centered AI, open-world recognition, particularly AI for healthcare (AI4Health), and cross-domain model generalization. He has published several papers such as CVPR, ACM MM, T-PAMI, T-FS, T-ASLP, PR, EAAI, and JBHI, etc.
\end{IEEEbiography}

\begin{IEEEbiography}[{\includegraphics[width=1in,height=1.25in,clip,keepaspectratio]{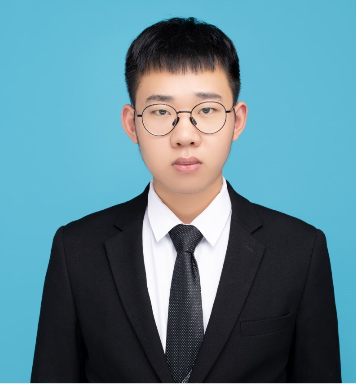}}]{Feng He} is pursuing a master's degree at the University of Science and Technology of China (USTC) in China. His main research interests lie in multimodal learning and makeup transfer. Prior to this, he graduated from the School of Computer Science at Yangtze University in June 2023.
\end{IEEEbiography}

\begin{IEEEbiography}[{\includegraphics[width=1in,height=1.25in,clip,keepaspectratio]{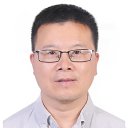}}]{Weijun Li} (Senior Member, IEEE) received the Ph.D. degree from the Institute of Semiconductors, Chinese Academy of Sciences (ISCAS), Beijing, China, in 2004.,He is currently a Professor of artificial intelligence with the Institute of Semiconductors, ISCAS and the University of Chinese Academy of Sciences, Beijing. He is an In Charge with the Artificial Intelligence Research Center, ISCAS and the Director of the Laboratory of Highspeed Circuits Neural Networks, ISCAS. His research interests include deep modeling, machine art, pattern recognition, artificial neural networks, and intelligent systems.
\end{IEEEbiography}

\begin{IEEEbiography}[{\includegraphics[width=1in,height=1.25in,clip,keepaspectratio]{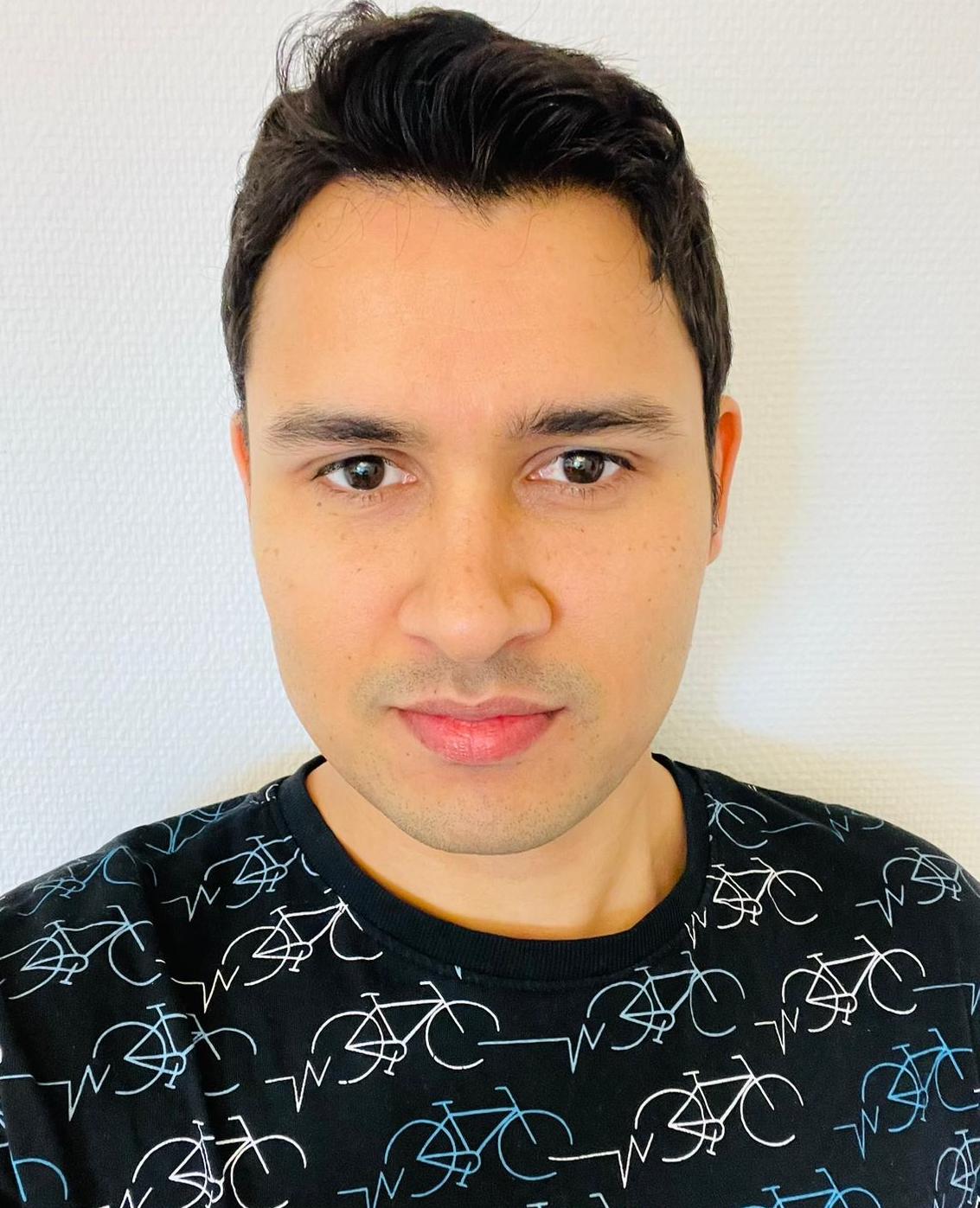}}]{Prayag Tiwari} (Senior Member, IEEE) received his Ph.D. degree from the University of Padova, Italy. He is currently working as a Senior Lecturer at Halmstad University, Sweden. Previously, he worked as a Postdoctoral Researcher at the Aalto University, Finland, and Marie Curie Researcher at the University of Padova, Italy. His name has appeared in the World’s Top 2\% of Scientists released by Stanford university and Elsevier. He received the 2022 Best Paper Award for a paper published in Neural Networks, Elsevier. His papers were also awarded as one of the Highly Cited Papers published in journals like applied sciences, measurements, etc. He has several publications in top journals and conferences, including Neural Networks, Journal of Physics A, Information Fusion, Knowledge-Based Systems, International Journal of Computer Vision, IEEE TNNLS, IEEE TFS, IEEE JBHI, IEEE TAI, IEEE IoTJ, IEEE BIBM, ACM TOIT, ACM CSUR, CIKM, SIGIR, AAAI, ECIR, ECML, ACL, etc. His research interests include Artificial Intelligence, Quantum Machine Learning, Deep Learning, Healthcare, Bioinformatics, NLP, Computer vision, Intelligent Transport, and IoT.
\end{IEEEbiography}

\begin{IEEEbiography}[{\includegraphics[width=1in,height=1.25in,clip,keepaspectratio]{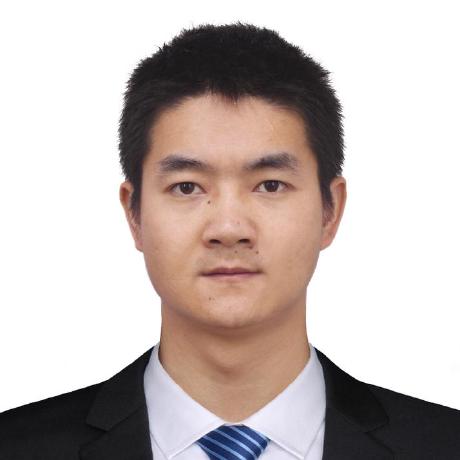}}]{Xinwang Liu} (Senior Member, IEEE) received the PhD degree from the National University of Defense Technology (NUDT), China. He is now professor with the School of Computer, NUDT. His current research interests include kernel learning and unsupervised feature learning. He has published more than 60 peer-reviewed papers, including those in highly regarded journals and conferences such as IEEE Transactions on Pattern Analysis and Machine Intelligence, IEEE Transactions on Knowledge and Data Engineering, IEEE Transactions on Image Processing, IEEE Transactions on Neural Networks and Learning Systems, IEEE Transactions on Multimedia, IEEE Transactions on Information Forensics and Security, ICML, NeurIPS, ICCV, CVPR, AAAI, IJCAI, etc. He serves as the associated editor of IEEE Transactions on Neural Networks and Learning Systems, IEEE Transactions on Cybernetics and Information Fusion Journal. More information can be found at https://xinwangliu.github.io/.
\end{IEEEbiography}

\end{document}